%% file: sample-sigconf.tex
\documentclass[sigconf]{acmart}
\usepackage{algorithm}
\usepackage{algpseudocode}
\usepackage{booktabs}
\usepackage{ragged2e}
\usepackage{enumitem}
\usepackage{multirow}
\definecolor{darkgrey}{rgb}{0.53,0.53,0.53}

\newcommand{\best}[1]{\textbf{#1}}
\newcommand{\second}[1]{\underline{#1}}
\usepackage{colortbl}
\definecolor{fbApp}{HTML}{ffe4e3}
\definecolor{tabhighlight}{HTML}{e5e5e5}

\usepackage{siunitx}
\usepackage{array}
\usepackage{threeparttable}
\usepackage{algorithm}
\usepackage{algpseudocode}
\usepackage{placeins} 
\usepackage{subcaption}
\newcolumntype{L}[1]{>{\raggedright\arraybackslash}p{#1}}
\usepackage{pifont}

\usepackage{afterpage}

\AtBeginDocument{%
  }

\copyrightyear{2025}
\acmYear{2025}
\setcopyright{cc}
\setcctype{by}
\acmConference[MM '25]{Proceedings of the 33rd ACM International Conference on Multimedia}{October 27--31, 2025}{Dublin, Ireland}
\acmBooktitle{Proceedings of the 33rd ACM International Conference on Multimedia (MM '25), October 27--31, 2025, Dublin, Ireland}\acmDOI{10.1145/3746027.3755458}
\acmISBN{979-8-4007-2035-2/2025/10}
\begin{document}

\title{DualSG: A Dual-Stream Explicit Semantic-Guided Multivariate Time Series Forecasting Framework}

\author{Kuiye Ding}
\email{dingkuiye@ict.ac.cn}
\affiliation{
  \institution{SKL of Processors, Institute of Computing Technology, Chinese Academy of Sciences}
  \city{Beijing}
  \country{China}
}

\author{Fanda Fan\textsuperscript{\dag}}
\thanks{\textsuperscript{\dag}Corresponding author.}
\email{fanfanda@ict.ac.cn}
\affiliation{
  \institution{SKL of Processors, Institute of Computing Technology, Chinese Academy of Sciences}
  \city{Beijing}
  \country{China}
}

\author{Yao Wang}
\email{yw5438@nyu.edu}
\affiliation{
  \institution{Fu Foundation School of Engineering and Applied Science, Columbia University}
  \city{New York}
  \country{USA}
}

\author{Ruijie Jian}
\email{jianruijie21@mails.ucas.ac.cn}
\affiliation{
  \institution{University of Chinese Academy of Sciences}
  \city{Beijing}
  \country{China}
}

\author{Xiaorui Wang}
\email{wangxiaorui20@mails.ucas.ac.cn}
\affiliation{
  \institution{University of Chinese Academy of Sciences}
  \city{Beijing}
  \country{China}
}

\author{Luqi Gong}
\email{luqi@zhejianglab.com}
\affiliation{
  \institution{Zhejiang Lab}
  \city{Hangzhou}
  \country{China}
}

\author{Yishan Jiang}
\email{22074104@emails.bjut.edu.cn}
\affiliation{
  \institution{Beijing University of Technology}
  \city{Beijing}
  \country{China}
}

\author{Chunjie Luo}
\email{luochunjie@ict.ac.cn}
\affiliation{
  \institution{Institute of Computing Technology, Chinese Academy of Sciences}
  \city{Beijing}
  \country{China}
}

\author{Jianfeng Zhan}
\email{zhanjianfeng@ict.ac.cn}
\affiliation{
  \institution{Institute of Computing Technology, Chinese Academy of Sciences}
  \city{Beijing}
  \country{China}
}
\affiliation{
  \institution{University of Chinese Academy of Sciences}
  \city{Beijing}
  \country{China}
}

\renewcommand{\shortauthors}{Kuiye Ding et al.}

\begin{abstract}
Multivariate Time Series Forecasting plays a key role in many applications. Recent works have explored using Large Language Models for MTSF to take advantage of their reasoning abilities. However, many methods treat LLMs as end-to-end forecasters, which often leads to a loss of numerical precision and forces LLMs to handle patterns beyond their intended design. Alternatively, methods that attempt to align textual and time series modalities within latent space frequently encounter alignment difficulty. In this paper, we propose to treat LLMs not as standalone forecasters, but as semantic guidance modules within a dual-stream framework. We propose \textbf{DualSG}, a \textbf{dual}-stream framework that provides explicit semantic guidance, where LLMs act as \textbf{S}emantic \textbf{G}uides to refine rather than replace traditional predictions. As part of DualSG, we introduce Time Series Caption, an explicit prompt format that summarizes trend patterns in natural language and provides interpretable context for LLMs, rather than relying on implicit alignment between text and time series in the latent space. We also design a caption-guided fusion module that explicitly models inter-variable relationships while reducing noise and computation. Experiments on real-world datasets from diverse domains show that DualSG consistently outperforms 15 state-of-the-art baselines, demonstrating the value of explicitly combining numerical forecasting with semantic guidance. The code is made available at \url{https://github.com/BenchCouncil/DualSG}
\end{abstract}




\maketitle

\section{Introduction}
Multivariate Time Series Forecasting (MTSF)~\cite{yu2024ginar,DSformer,HybridZheng,DBLP:conf/aaai/HuangSZCDZW24} plays an essential role in a wide range of real-world applications, including health monitoring, weather prediction, and financial decision-making~\cite{peng2023, wu2023, zhu2024, bi2023, van2023}. Traditionally, most forecasting models rely on a single modality, using either statistical methods or deep learning architectures~\cite{brain1997, qiuTSF, timesnet, miao2024, shao2024exploring}. Recently, a growing number of MTSF methods have explored incorporating multimodal information by using Large Language Models (LLMs)~\cite{timemmd, ChatTime, timecma, gpt4ms, timellm}, which has sparked both interest and controversy~\cite{tan2024are}. While LLMs have demonstrated strong reasoning abilities in various domains, their effectiveness in multivariate time series forecasting under question. To better understand this issue, we first review two predominant paradigms of current LLM-based forecasting methods.

\begin{figure}[t!]
    \centering
    \includegraphics[width=0.48\textwidth]{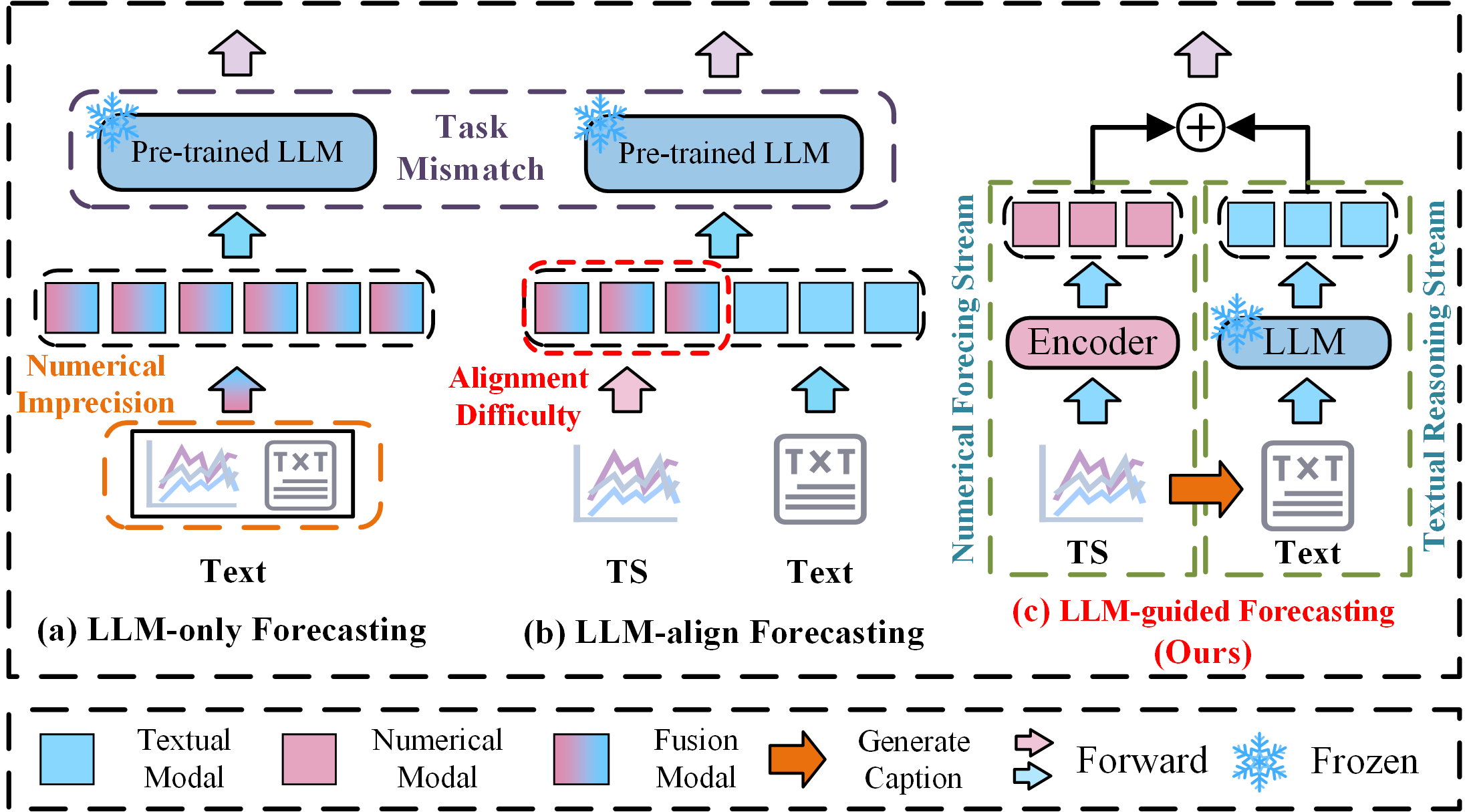}
    \caption{The LLM-based paradigms for MTSF. (a) LLM-only Forecasting: treat time series as natural language for MTSF. (b) LLM-align Forecasting: alignment difficulty between modalities. (c) Dual-Stream LLM: numerical forecasting stream and textual reasoning stream.}
    \label{fig:idea} 
\end{figure}

Many LLM-based forecasting methods can be broadly categorized into two types. \textbf{LLM-only Forecasting Paradigm}~\cite{llmtime, tsfl, liu2024autotimes, ChatTime} treat time series as text by converting them into natural language, and perform forecasting purely via LLMs. This design often leads to \textbf{numerical imprecision}, as continuous values must be discretized or approximated in text. It also leads to a \textbf{task mismatch}, as LLMs, which are inherently designed for natural language processing, are being misapplied to model time series data. Due to the limited output token budget of LLMs, this often restricts the feasible forecasting horizon. \textbf{LLM-align Forecasting Paradigm}~\cite{timellm, timecma, test, s2ip} align textual and time series modalities at the token level, as illustrated in Figure~\ref{fig:idea}(b). This approach frequently suffers from alignment difficulty, which distorts key time series properties, ultimately compromising the integrity of essential forecasting signals. Both paradigms thus exhibit significant shortcomings in numerical precision and alignment difficulty, raising fundamental doubts about the suitability of LLMs as forecasters for time series.

\begin{table}[t!]
    \scriptsize
    \centering
    \caption{Examples of different prompt types used in LLM-based time series forecasting.}
    \label{tab:three}
    \rowcolors{2}{gray!10}{white}
    \begin{tabular}{p{0.45\columnwidth} p{0.45\columnwidth}}
        \toprule
        \textbf{Type} & \textbf{Prompt Example} \\
        \midrule
        LLM-only & From August 16, 2019, Friday to August 30, 2019, Friday, the average temperature of region 110 was 78, 81, 83, 84, 84, 82, 83, 78, 77, 77, 74, 77, 78, 73, 76 degrees on each day. \\
        LLM-align & "rise", "up", "down", "steady", "top" \\
        \textbf{LLM-guided (Ours)}& The time series exhibits a gradual decrease from start to end, with a flat end. \\
        \bottomrule
    \end{tabular}
\end{table}


To address these challenges, we propose a new paradigm: the dual-stream LLM framework, which treats LLMs not as standalone forecasting models, but as semantic reasoning components, as illustrated in Figure~\ref{fig:idea}(c).  See example in Table \ref{tab:three}. This framework combines a numerical forecasting stream with a textual reasoning stream, enabling complementary modeling of time series dynamics. The former focuses on fine-grained temporal modeling, while the latter offers coarse-grained semantic guidance. Together, they aim to mitigate the numerical imprecision, task mismatch, and modality alignment difficulties observed in previous approaches.

To implement this dual-stream framework, we design two complementary components: a numerical forecasting stream for fine-grained signal modeling, and a textual reasoning stream for semantic-level correction. Together, these two streams address the core challenges identified in prior LLM-based approaches, numerical imprecision, task mismatch, and modality alignment difficulty~\cite{llmtime, timecma, timellm, ChatTime}. 1) The \textbf{numerical forecasting stream} captures multi-scale temporal patterns through a Multi-scale Adaptive Patching (\textbf{MAP}) mechanism, which encodes the raw multivariate series into hierarchical representations across different resolutions. This enables more precise modeling of complex dynamics and reduces the risk of signal distortion, directly mitigating the numerical precision loss observed in LLM-only designs~\cite{llmtime, tsfl}. 2) The \textbf{textual reasoning stream}, in turn, uses \textbf{Time Series Captions (TSCs)} as trend-level summaries to guide semantic correction. Unlike previous methods that perform alignment implicitly in latent space and at the token level~\cite{timellm, timecma, s2ip, llmps, hu2025contextalignment}, we adopt \emph{explicit alignment} at the sentence level. These two streams are complementary: the numerical stream excels at modeling local patterns and short-term variations, but often lacks awareness of broader temporal trends. The textual stream provides high-level semantic correction to compensate for this limitation, helping the model maintain consistency and accuracy over longer horizons.

While the dual streams in DualSG capture complementary aspects of time series, modeling variable-wise dependencies remains challenging. Existing methods, including full inter-channel attention~\cite{zhang2023crossformer, shi2025timemoe}, channel-wise independence~\cite{liu2024itransformer, patchtst}, and clustering-based grouping~\cite{qiu2025duet, liu2024dgcformerdeepgraphclustering}, involve trade-offs among scalability, expressiveness, and adaptability, and often lack semantic reasoning. To address this, we propose \textbf{SemFuse}, a lightweight fusion module that leverages TSCs to guide sparse and interpretable inter-channel interactions via trend-level semantic similarity. To integrate the numerical and textual streams, we introduce the \textbf{Spatial \& Temporal Attention Matrix (STAM)}, which dynamically adjusts their contributions across time and variables. Instead of uniform fusion, STAM performs context-aware weighting to emphasize either fine-grained signals or semantic corrections as needed, enhancing both short-term accuracy and long-term trend consistency under complex or distribution-shifted scenarios.

\textbf{Our main contributions are as follows:}
\begin{itemize}
    \item We propose \textbf{DualSG}, a dual-stream framework that combines numerical forecasting and textual reasoning, addressing the limitations of LLM-only and LLM-align paradigms.

    \item We introduce \textbf{Time Series Caption (TSC)} as interpretable, trend-aware prompts that explicitly align LLM reasoning with forecasting tasks and enable semantic correction of numerical outputs.

    \item We design two semantic-aware fusion modules: \textbf{SemFuse}, which captures inter-channel dependencies via caption-level relevance, and \textbf{STAM}, which adaptively fuses the two streams based on spatio-temporal context.

\end{itemize}





\section{Related Work}
\subsection{Time Series Caption}

Time Series Caption~\cite{jhamtani-berg-kirkpatrick-2021-truth, lincap, murakami-etal-2017-learning, sumts, trabelsi2025timeserieslanguagemodel} focuses on generating natural language descriptions that summarize key patterns in temporal data. Early work such as TRUCE~\cite{jhamtani-berg-kirkpatrick-2021-truth} used modular neural programs to produce structured summaries, but relied heavily on hand-crafted rules and lacked scalability. Later methods such as TSLM~\cite{trabelsi2025timeserieslanguagemodel} introduced subtitle-style captions to address data limitations, while TESSA~\cite{lin2024decodingtimeseriesllms} explored multi-agent systems for domain-specific annotation. Despite these efforts, the field remains underdeveloped due to the lack of large-scale datasets and open benchmarks.

While prior work has explored time series caption for purposes such as narrative generation or explainability, we are the first to leverage TSC as a form of semantic guidance for time series forecasting. Specifically, we introduce a Time Series Caption Generation (TSCG) module tailored for this task. Rather than focusing on narrative quality, our TSCG module produces trend-aware, high-level summaries that align with LLM capabilities and facilitate both forecast correction and cross-modal integration within our dual-stream architecture.
\subsection{Multivariate Time Series Forecasting}

\subsubsection{Traditional Models}

Multivariate time series forecasting has traditionally relied on statistical models such as ARIMA~\cite{box1970distribution} and STL~\cite{cleveland1990stl}, which capture trend and seasonality through explicit decomposition. With the development of deep learning, models based on RNNs~\cite{10.1145/3292500.3330672}, CNNs~\cite{timesnet, wang2023micn}, Transformers~\cite{liu2024itransformer, patchtst}, and MLPs~\cite{das2023longterm,bao2024boundaryaware, dlinear, zhang2024not} have become widely used for their ability to model sequential dependencies and long-range interactions. 

Despite architectural advances, these models remain fully numerical and deterministic, with limited generalization across domains and poor adaptability to distribution shift and temporal variability~\cite{timellm}. They also offer limited interpretability, especially in scenarios that require semantic understanding or external context.

\subsubsection{LLM-based Forecasting Methods}
Recent methods have explored enhancing time series forecasting by incorporating Large Language Models (LLMs), either through tokenizing numerical sequences~\cite{llmtime,merrill}, adding domain-specific prompts~\cite{gpt4ts,timellm,liu2024unitime}, training time-aware language models from scratch~\cite{liu2024timeffm,das2024decoderonlyfoundationmodeltimeseries}, or integrating multimodal data such as charts or quantized representations~\cite{meng-etal-2024-chartassistant,calf,lee2023vectorquantizedtimeseries}. 

Despite promising results, these approaches face recurring issues: frozen LLMs introduce token inefficiency and numerical imprecision~\cite{merrill}, hybrid models suffer from alignment gaps between modalities~\cite{timellm,liu2024unitime}, and end-to-end or codebook-based methods often lose semantic detail or require costly data formats~\cite{calf,lee2023vectorquantizedtimeseries}.

Our work differs in two key aspects:  
(1) Instead of relying on low-level statistics or free-form text, we introduce coarse-grained semantic abstraction via \textit{time series captions}, which describe global trends in natural language;  
(2) Rather than aligning modalities through tokenization or projection, we design a dual-stream forecasting framework that enables flexible and interpretable interaction between the numerical forecasting stream and the textual reasoning stream.

\begin{figure*}[t!]
    \centering
    \includegraphics[width=1\textwidth]{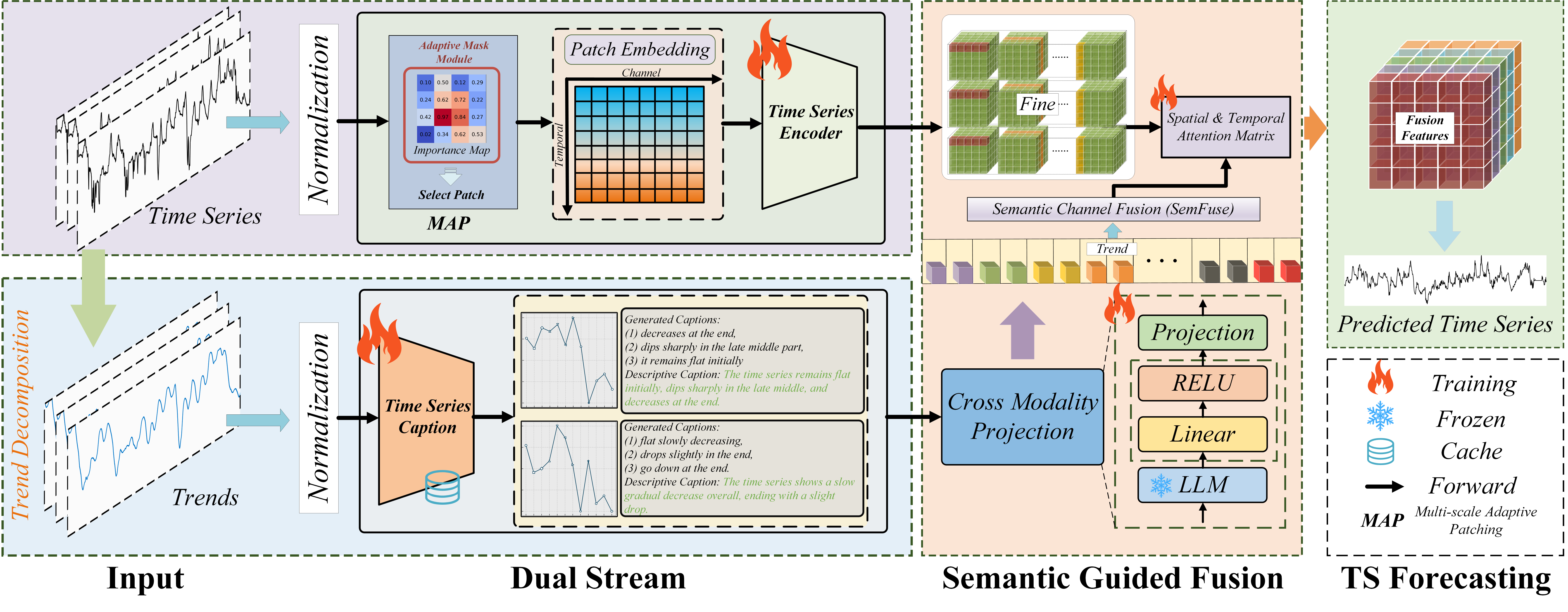}
    \caption{
    Overview of the \textbf{DualSG} framework.
    (a) The \textit{numerical forecasting stream} applies MAP with multi-scale receptive fields to extract temporal features.
    (b) The \textit{textual reasoning stream} generates TSCs through trend decomposition and caption generation, then performs semantic reasoning via LLM-based trend correction.
    (c) \textit{SemFuse} builds sparse semantic links between channels using caption-guided attention masking.
    (d) A dynamic fusion module combines numerical forecasts with semantic trend-level corrections.
    }
    \label{fig:framwork}
\end{figure*}

\section{Methodology}

\subsection{Numerical Forecasting Stream}

We design the Numerical Forecasting Stream to be lightweight yet expressive, focusing on fine-grained temporal modeling without relying on external inputs. Unlike traditional approaches that apply uniform patching, we adopt adaptive multi-scale segmentation to better capture temporal patterns while reducing overfitting risks.

A key challenge lies in the tokenization of time series for transformer-based models. Most existing methods use fixed-size patches, which differ fundamentally from subword-based tokenization in NLP (e.g., BPE~\cite{bostrom2020bytepairencodingsuboptimal}). Such static segmentation neglects the varying temporal relevance of different regions. Recent data often benefits short-term prediction, whereas earlier patterns inform long-term trends~\cite{WANG2023110214}.

This discrepancy between coarse, fixed patches in time series and dynamic, frequency-aware segments in text can impair cross-modal alignment. In particular, oversized patches may dominate attention and suppress fine-grained semantic cues, thereby weakening the integration of numerical and textual modalities.

\vspace{0.5em}
To address this, we propose \textbf{Multi-scale Adaptive Patching (MAP)}, a non-uniform segmentation method that creates multi-scale temporal tokens. To our knowledge, no prior patch-based time series methods~\cite{chen2024pathformer, wang2024medformer, kudrat2025patchwisestructurallosstime} explicitly consider temporal decay effects in patch design.

\vspace{0.75em}
\textbf{Temporal Partitioning.}  
Given a sequence \( X \in \mathbb{R}^T \), we divide it into recent, mid-range, and distant regions using learnable quantile thresholds \( q_1, q_2 \in \mathbb{R} \), dynamically adjusted by the prediction length \( \tau \):
\begin{equation}
\begin{aligned}
\mathcal{R}_{\text{near}} &= [T - q_1, T], \quad
\mathcal{R}_{\text{mid}}  = [T - q_2, T - q_1), \\
\mathcal{R}_{\text{far}}  &= [1, T - q_2)
\end{aligned}
\end{equation}
\begin{equation}
(q_1, q_2) = f_{\text{quantile}}(\tau; \theta_q)
\end{equation}
where \( f_{\text{quantile}} \) is a two-layer MLP with sigmoid activations. This adaptive design eliminates manual thresholding and adjusts to varying horizons.

\vspace{0.75em}
\textbf{Multi-scale Patch Generation.}  
Each region \( \mathcal{R}_i \) is segmented using region-specific patch size \( s_i \) and stride \( \lfloor s_i/2 \rfloor \):
\begin{equation}
\mathcal{P}_i = \{X_{t:t+s_i} \mid t \in \mathcal{R}_i, \ t \equiv 0 \ (\text{mod}\ \lfloor s_i/2 \rfloor)\}
\end{equation}
Patches are projected via a region-specific MLP:
\begin{equation}
E_i = \text{GeLU}(\text{MLP}(\mathcal{P}_i; \theta_i)) \in \mathbb{R}^{|\mathcal{P}_i| \times d}
\end{equation}
This yields region-aware embeddings for subsequent filtering.

\vspace{0.75em}
\textbf{Adaptive Importance Masking.}  
To estimate token importance, we apply a gradient-based scoring function:
\begin{equation}
\alpha_i = \text{Sigmoid}\left( 
\frac{\|\nabla_t E_i\|_2}{\max(\|\nabla_t E_i\|_2)} \cdot \text{MSA}(E_i) W_\alpha 
\right)
\end{equation}
We select top-\(k_i\) patches via soft top-k sampling:
\begin{equation}
\mathcal{M}_i = \text{SoftTopk}(\alpha_i, \ k_i = \lceil \rho_i |\mathcal{P}_i| \rceil), \quad
\tilde{E}_i = \mathcal{M}_i \odot \mathcal{P}_i
\end{equation}
The \(\text{SoftTopk}(\cdot)\) ensures differentiability, and total selection is constrained to match uniform segmentation for efficiency.

\subsection{Textual Reasoning Stream}

The textual reasoning stream is designed to extract high-level semantic patterns from raw time series inputs. To the best of our knowledge, this is the first work that leverages time series captions to directly enhance forecasting performance. At the core of this component lies the \textbf{Time Series Caption Generation (TSCG)} module, which converts time series segments into natural language summaries. These generated captions serve as semantic prompts that guide the LLM to refine trend-level predictions.

Due to the lack of publicly available models for time series caption generatation, we design a lightweight encoder–decoder architecture as the TSCG module. It is pre-trained on large-scale mixed-domain time series data to learn to describe trend-level dynamics in natural language. The TSCG module, adopts an autoregressive encoder–decoder architecture with a built-in cross-modal alignment mechanism.

\vspace{0.5em}
\textbf{Patch Embedding.}  
Given an input sequence \( X \in \mathbb{R}^{T} \), we divide it into overlapping patches using a fixed patch length \( P \) and stride \( S \). This results in \( N = \left\lfloor \frac{T - P}{S} \right\rfloor + 2 \) overlapping patches, denoted as \( X_{\text{p}} \in \mathbb{R}^{N \times P} \).

Each patch is linearly projected into a \( d \)-dimensional embedding space and enriched with positional encoding:
\begin{equation}
H_p = X_{\text{p}} W_p + X_{\text{pos}}, \quad W_p \in \mathbb{R}^{P \times d}, \quad X_{\text{pos}} \in \mathbb{R}^{N \times d}
\end{equation}
Here, \( H_p \in \mathbb{R}^{N \times d} \) is the patch embedding matrix.

\vspace{0.5em}
\textbf{Temporal Encoding.}  
The embeddings \( H_p \) are processed by  a \( K \)-layer Transformer encoder to capture hierarchical temporal dependencies:
\begin{align}
H' &= \text{MSA}(\text{LayerNorm}(H_p)) + H_p \\
H_{\text{enc}} &= \text{FFN}(\text{LayerNorm}(H')) + H' \\
H_{\text{enc}}^{(k)} &= \text{TransformerBlock}(H_{\text{enc}}^{(k-1)}), \quad k = 1, \dots, K
\end{align}

\vspace{0.5em}
\textbf{Time-to-Text Self Attention.}  
To generate natural language captions directly from time series, we employ a multi-head self-attention mechanism that maps temporal representations into textual space. Different from standard encoder-decoder cross attention, our design does not rely on external textual queries; instead, the decoder operates solely on time-series features.  

Formally, given the encoded time series \( H_{\text{time}} \in \mathbb{R}^{T \times d} \), we compute the attention states as:  
\begin{equation}
Q = H_{\text{time}} W_Q, \quad K = H_{\text{time}} W_K, \quad V = H_{\text{time}} W_V
\end{equation}
\begin{equation}
H_{\text{attn}} = \text{softmax}\left(\frac{QK^\top}{\sqrt{d}}\right)V
\end{equation}
\begin{equation}
H_{\text{align}} = \text{LayerNorm}(H_{\text{time}} + H_{\text{attn}})
\end{equation}

This formulation directly injects temporal signals into the GPT2 decoder, allowing it to gradually transform time-series dynamics into linguistic representations. By avoiding mixed-modal cross-attention, the model ensures that the textual output is conditioned purely on numerical patterns, while still benefiting from the flexibility of pre-trained language models.

\vspace{0.5em}
\textbf{Caption Decoding.}  
The decoder generates captions in an autoregressive fashion. At each step \( t \), it consumes previously generated tokens \( y_{<t} \) and the aligned encoder output to predict the next token:
\begin{align}
E_{\text{dec}}^t &= \text{Embed}(y_{<t}) + E_{\text{pos}}^t \\
H_{\text{dec}}^t &= \text{TransformerDecoder}(E_{\text{dec}}^t, H_{\text{align}}) \\
p(y_t | y_{<t}, X) &= \text{softmax}(W_o H_{\text{dec}}^t)
\end{align}
where \( W_o \in \mathbb{R}^{d \times |\mathcal{V}|} \) maps decoder outputs to vocabulary logits.

\vspace{0.5em}
\textbf{Training and Inference.}  
The captioning model is trained using a cross-entropy loss, which encourages the decoder to predict each ground-truth token conditioned on its preceding tokens:
\begin{equation}
\mathcal{L}_{\text{TSCG}} = -\sum_{t=1}^{M} \log p(y_t | y_{<t}, X)
\end{equation}
During inference, the decoder generates tokens sequentially until an end-of-sequence marker is emitted. This process enables zero-shot captioning for previously unseen temporal patterns. These captions are specifically designed to guide the numerical stream by correcting trend-level forecasting errors, thereby addressing the challenge of \textbf{modality misalignment} between textual and numerical representations.

\textbf{Trend Semantics.} We define TSCs as structured natural language sequences that abstract the coarse-grained behavior of time series segments. Each caption encodes trend-level semantics across three dimensions: direction (e.g., rising, falling), intensity (e.g., sharp, gradual), and temporal transition (e.g., followed by, then). This structured abstraction serves as a semantic scaffold, helping the reasoning stream to align numerical forecasts with interpretable trend patterns and correct high-level prediction drifts. In this sense, TSC provides a controllable, interpretable form of semantic regularization for long-horizon forecasting. 


\subsection{Dual-Stream Fusion}

To integrate the outputs of the two streams in a principled and efficient manner, we propose a dual-stage fusion strategy. The first stage leverages semantic information for channel-level interaction through \textbf{SemFuse}. The second stage performs dynamic fusion between the numerical and textual streams using a learnable attention mechanism. Together, these components enhance forecasting performance by aligning precise numerical patterns with high-level semantic trends.

\vspace{0.75em}
\subsubsection{Channel Fusion via SemFuse}

The \textbf{SemFuse} module is positioned after the Textual Reasoning Stream. It uses the generated captions as semantic anchors to construct interpretable inter-channel dependencies. As illustrated in Figure~\ref{fig:framwork}(c), the module performs sparse fusion in three stages: semantic relevance scoring, sparse top-k selection, and gated feature aggregation. For example, if two variables are both described by captions such as “rising together and then dropping,” SemFuse will identify them as semantically aligned and enable them to exchange trend-level information.
 

\textbf{Stage 1: Semantic Relevance Projection.}
For each pair of channels \( n, i \), we compute their semantic affinity using a frozen LLM that encodes the caption \( S_n \) for channel \( n \):
\begin{equation}
g_{n,i} = \operatorname{Sigmoid} \left( \mathbf{W}_{\text{proj}} \cdot \operatorname{LLM}(S_n) \right)[i]
\end{equation}
Here, \( g_{n,i} \in [0,1] \) quantifies the relevance between channels \( n \) and \( i \) based on caption semantics. \(\mathbf{W}_{\text{proj}} \in \mathbb{R}^{N \times d}\) is a learnable projection that maps the LLM-encoded caption into an $N$-dimensional affinity space.

\textbf{Stage 2: Sparse Top-K Selection.}
To promote sparsity and reduce redundancy, we construct a binary mask \( \mathcal{M}_{n,i} \in \{0,1\} \) by selecting the top-\( K \) most relevant channels:
\begin{equation}
\mathcal{M}_{n,i} =
\begin{cases}
1, & g_{n,i} \in \text{TopK}(\{g_{n,j}\}_{j=1}^N, K) \\
0, & \text{otherwise}
\end{cases}
\end{equation}

\textbf{Stage 3: Gated Channel Fusion.}
Each channel's final fused representation is computed by aggregating the selected channels, weighted by their semantic scores:
\begin{equation}
F_n^{\text{fused}} = \operatorname{GeLU} \left( \mathbf{W}_{\text{fuse}} \cdot \text{Concat} \left( F_n, \sum_{i=1}^N \mathcal{M}_{n,i} \cdot g_{n,i} \cdot F_i \right) \right)
\end{equation}
where \( F_i \in \mathbb{R}^d \) is the feature of channel \( i \), and \( \mathbf{W}_{\text{fuse}} \) is a shared projection matrix.

\textbf{Efficiency and Robustness.}  
Compared to Channel Dependency (CD) methods that compute full pairwise attention, and Channel Clustering (CC) approaches that enforce static groupings, SemFuse is more efficient and flexible. By fusing channels through a one-shot, caption-guided pathway, SemFuse avoids exhaustive computation while retaining interpretability. Since fusion occurs at the semantic level, it is also more robust to local fluctuations in raw inputs. 



\vspace{0.75em}
\subsubsection{Collaborative Forecasting with Cross-Stream Fusion}

The second fusion stage integrates the outputs of the numerical and textual streams to generate a unified forecast. The \textbf{Numerical Forecasting Stream} captures fine-grained patterns from raw sequences, while the \textbf{Textual Reasoning Stream} provides coarse trend correction derived from captions.

\textbf{Numerical Forecasting Stream.}
Given an input \( X \in \mathbb{R}^{N \times T} \), we first apply MAP to extract multi-scale temporal patches \( \mathcal{P} \). These are fed into a temporal encoder to generate the base forecast:
\begin{equation}
\hat{Y}_{\text{base}} = \text{TemporalEncoder}(\mathcal{P})
\end{equation}

\textbf{Textual Reasoning Stream.}
Captions \( \mathcal{S} \) are generated using the TSCG module, and then encoded via a frozen LLM. The output is projected into a semantic correction term \( \Delta \hat{Y} \), represented as a piecewise-constant sequence:
\begin{equation}
\Delta \hat{Y} = \text{Repeat}\left( \text{LayerNorm} \left( \text{ReLU}(W \cdot \text{LLM}(\mathcal{S}) + b) \right) \right)
\end{equation}
Here, \texttt{Repeat} expands each semantic unit over a fixed interval, enabling implicit periodicity modeling.

\textbf{Dynamic Stream Fusion.}
A learnable attention weight \( \mathcal{W}_{\text{STAM}} \in [0,1] \) adaptively balances the contributions of the two streams:
\begin{equation}
\hat{Y} = \mathcal{W}_{\text{STAM}} \odot \hat{Y}_{\text{base}} + (1 - \mathcal{W}_{\text{STAM}}) \odot \Delta \hat{Y}
\end{equation}

This dual-stream fusion preserves numerical precision while incorporating semantic corrections to mitigate trend-level biases. The joint prediction benefits from both localized signals and abstract reasoning, improving robustness under distribution shifts. This explicit separation allows each stream to specialize, mitigating \textbf{numerical imprecision} and \textbf{task mismatch} introduced by relying solely on LLMs.

\section{Experiments}
\subsection{Experimental Settings}
\subsubsection{Datasets}

We evaluate the performance of \textbf{DualSG} and baseline models on 12 widely-used multivariate time series datasets from the Time Series Library~\cite{timesnet}. These include benchmark datasets such as \textit{Weather}, \textit{Traffic}, \textit{Electricity}, \textit{Solar-Energy}, and four \texttt{ETT} variants (ETTh1, ETTh2, ETTm1, ETTm2). 


\subsubsection{Baselines}

We choose the latest state-of-the-art models to serve as baselines, including CNN-based models~(TimesNet~\cite{timesnet}, MICN~\cite{wang2023micn}), MLP-based models (DLinear~\cite{dlinear} and TimeMixer~\cite{wang2024timemixer}), Transformer-based models~(PatchTST~\cite{patchtst}, iTransformer~\cite{liu2024itransformer}, FEDformer~\cite{zhou2022fedformer}, Crossformer~\cite{zhang2023crossformer}, and Non-stationary Transformer (Stationary)~\cite{liu2022non}), and LLM-based models~(CALF~\cite{calf}, TimeLLM~\cite{timellm}, GPT4TS~\cite{gpt4ts}). The Fusion Module in the DUET~\cite{qiu2025duet} model borrows the design concept of Transformer, so this work categorizes it as a transformer-based.

\subsubsection{Implementation Details}

In the DualSG training phase, the parameters of the TSCG module are kept frozen, and the module is employed exclusively to generate captions through inference. All models are implemented in PyTorch 2.0.1 and trained using the Adam optimizer with cosine annealing learning rate scheduling. We use L1 loss as the objective function. Experiments are conducted on four NVIDIA A800 GPUs (80GB each). The batch size is initialized at 32 and dynamically adjusted to avoid out-of-memory errors using gradient accumulation and memory-efficient attention. We report performance using standard metrics: \textbf{Mean Squared Error (MSE)} and \textbf{Mean Absolute Error (MAE)}.

\input{table/ablation}

\begin{figure}[h]
    \centering
    \includegraphics[width=0.4\textwidth]{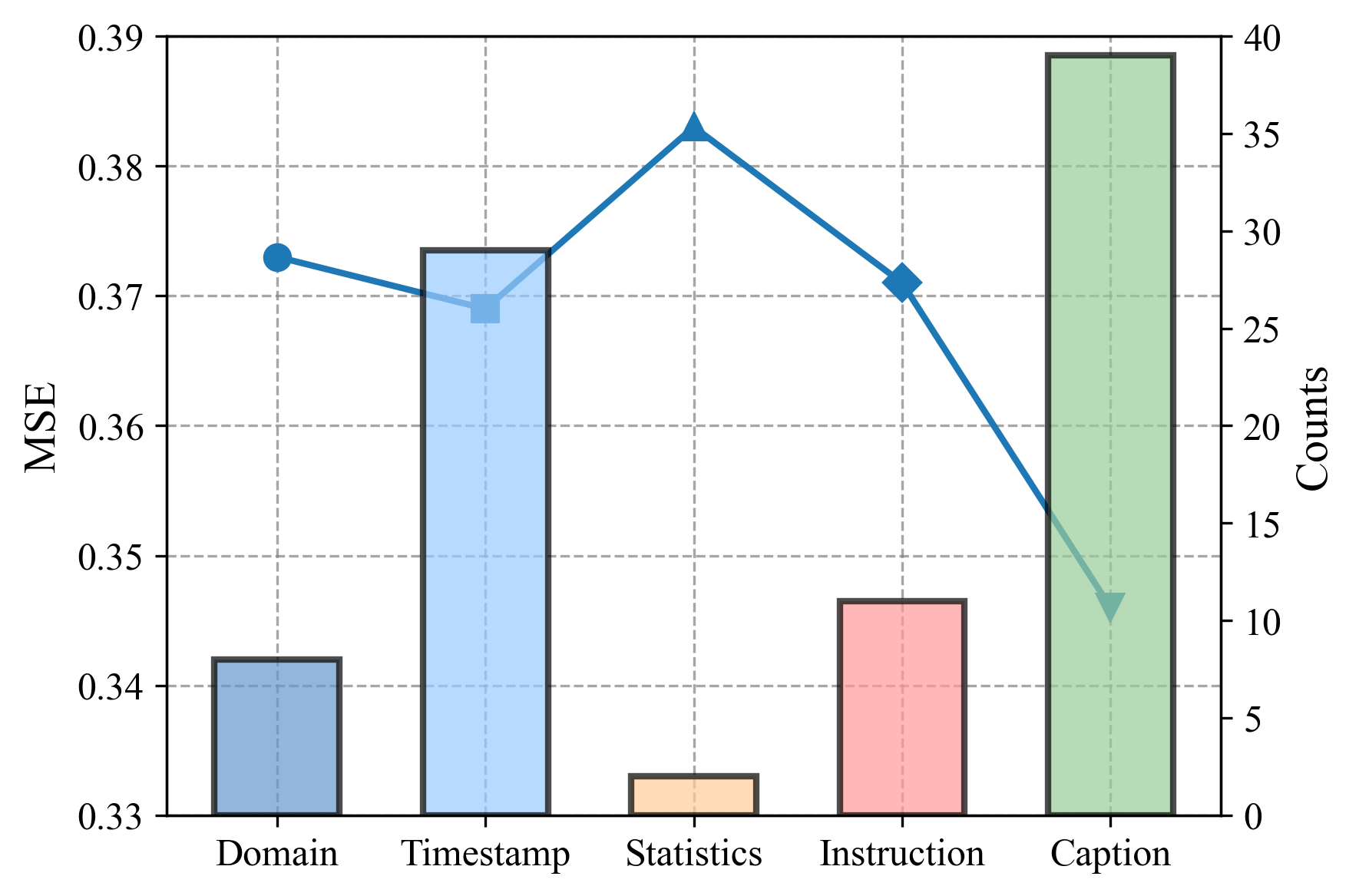}
    \caption{Ablation study comparing various prompt strategies for MTSF. The bar chart refers to the counts of best outcomes for different prompts; The line graphs illustrate the average MSE for various prompts across multiple datasets.}
    \label{fig:prompt}
\end{figure}

\subsection{Main Results}

Comprehensive forecasting results are listed in Table~\ref{tab::long-term} with the best in bold and the second underlined. We summarize three key observations:

\paragraph{1) Superior Performance Across Architectures}
DualSG achieves state-of-the-art results across both LLM-based and non-LLM models in long-term forecasting (Table~\ref{tab::long-term}). Compared to the second-best LLM baseline (CALF), it reduces MSE by 11.3\% and MAE by 9.8\% on average across eight datasets. On Solar-Energy, it lowers MSE by 40.7\% (0.192 vs. 0.324), demonstrating effective semantic-guided channel fusion. It also consistently outperforms the strongest non-LLM baseline (DUET).

\paragraph{2) Robust Channel Interaction via Semantic Fusion}
SemFuse enhances cross-channel modeling by leveraging semantic relevance to suppress noise. On ETTm2, where inter-channel correlation is weak, DualSG achieves an MSE of 0.254, outperforming Crossformer (1.216, 79.1\% reduction) and PatchTST (0.285, 10.9\% reduction), confirming its ability to avoid overfitting to spurious dependencies.

\paragraph{3) Collaborative Forecasting}
DualSG excels on non-stationary datasets by leveraging dual-stream collaboration. Compared to FEDformer, which targets non-stationary series, DualSG reduces MSE by 35.1\% on Traffic and 26.2\% on Weather, mainly benefiting from high-level trend correction via the semantic stream.

\begin{figure}[ht]
    \vspace{-3mm}
    \centering
    \begin{subfigure}[t]{0.205\textwidth}
        \includegraphics[width=\linewidth]{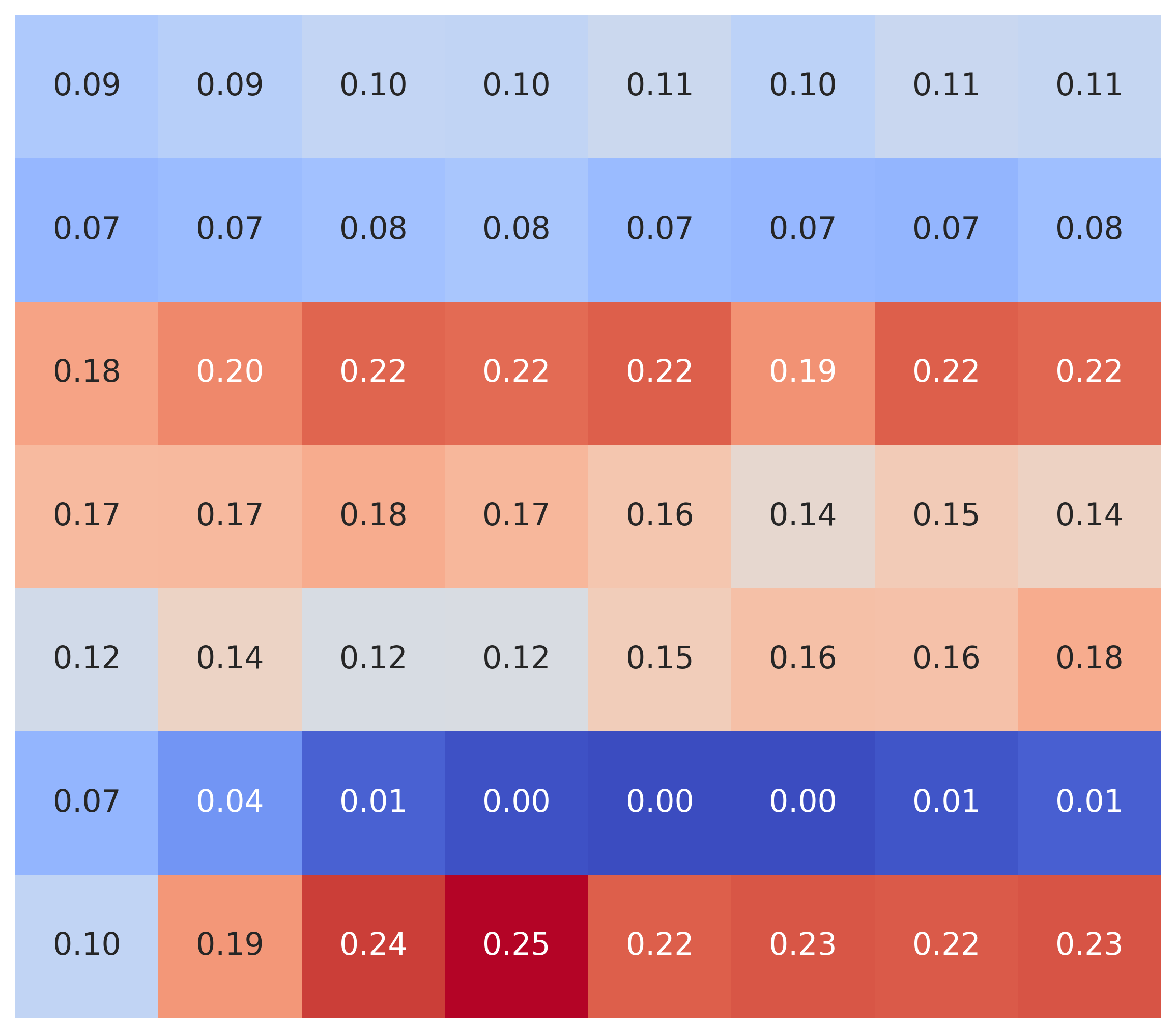}
        \vspace{-5mm}
        \caption{\scriptsize 96 Predicted Length}
    \end{subfigure}
    \hfill
    \begin{subfigure}[t]{0.205\textwidth}
        \includegraphics[width=\linewidth]{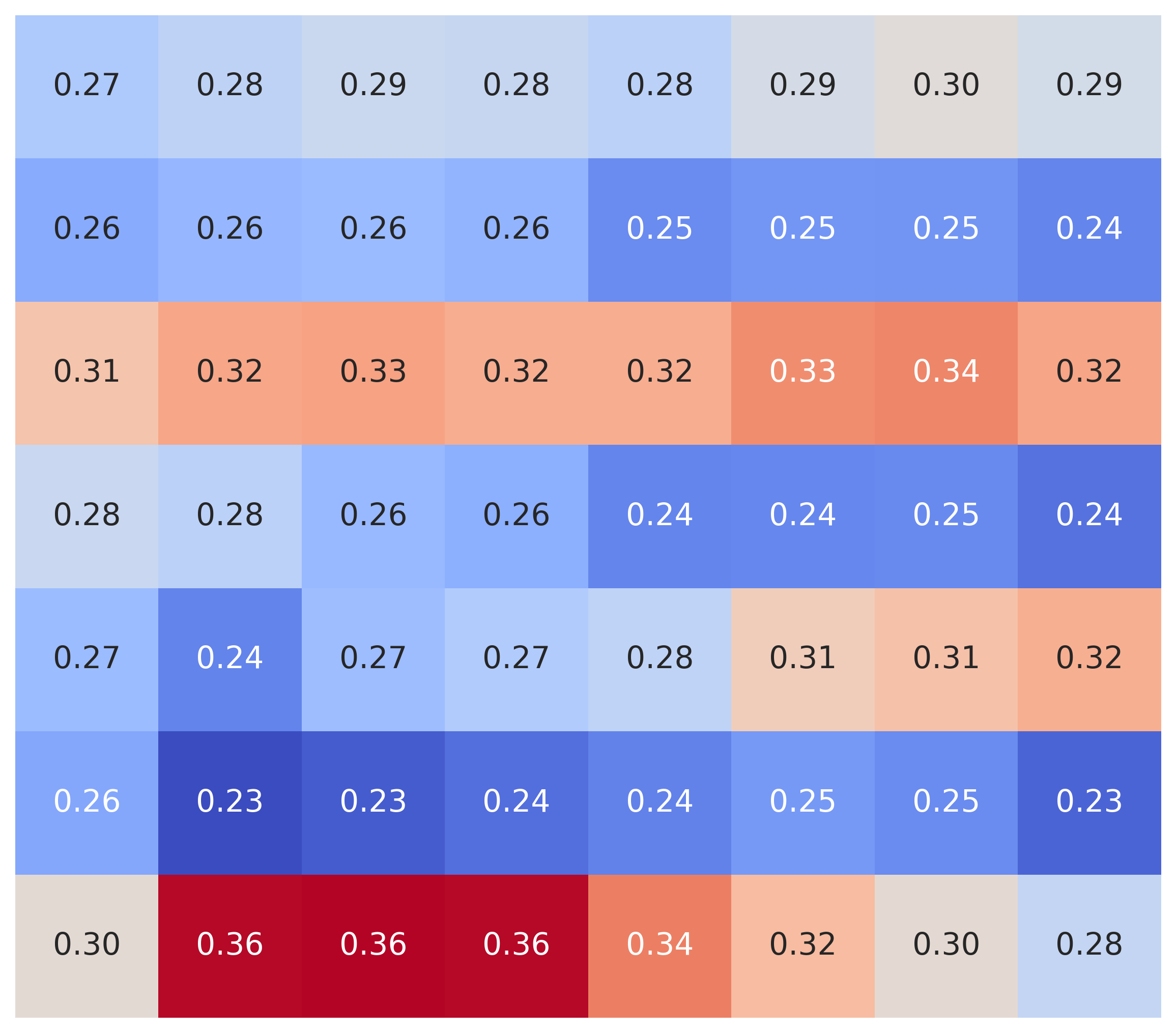}
        \vspace{-5mm}
        \caption{\scriptsize 192 Predicted Length}
    \end{subfigure}
    \hfill
    \begin{subfigure}[t]{0.205\textwidth}
        \includegraphics[width=\linewidth]{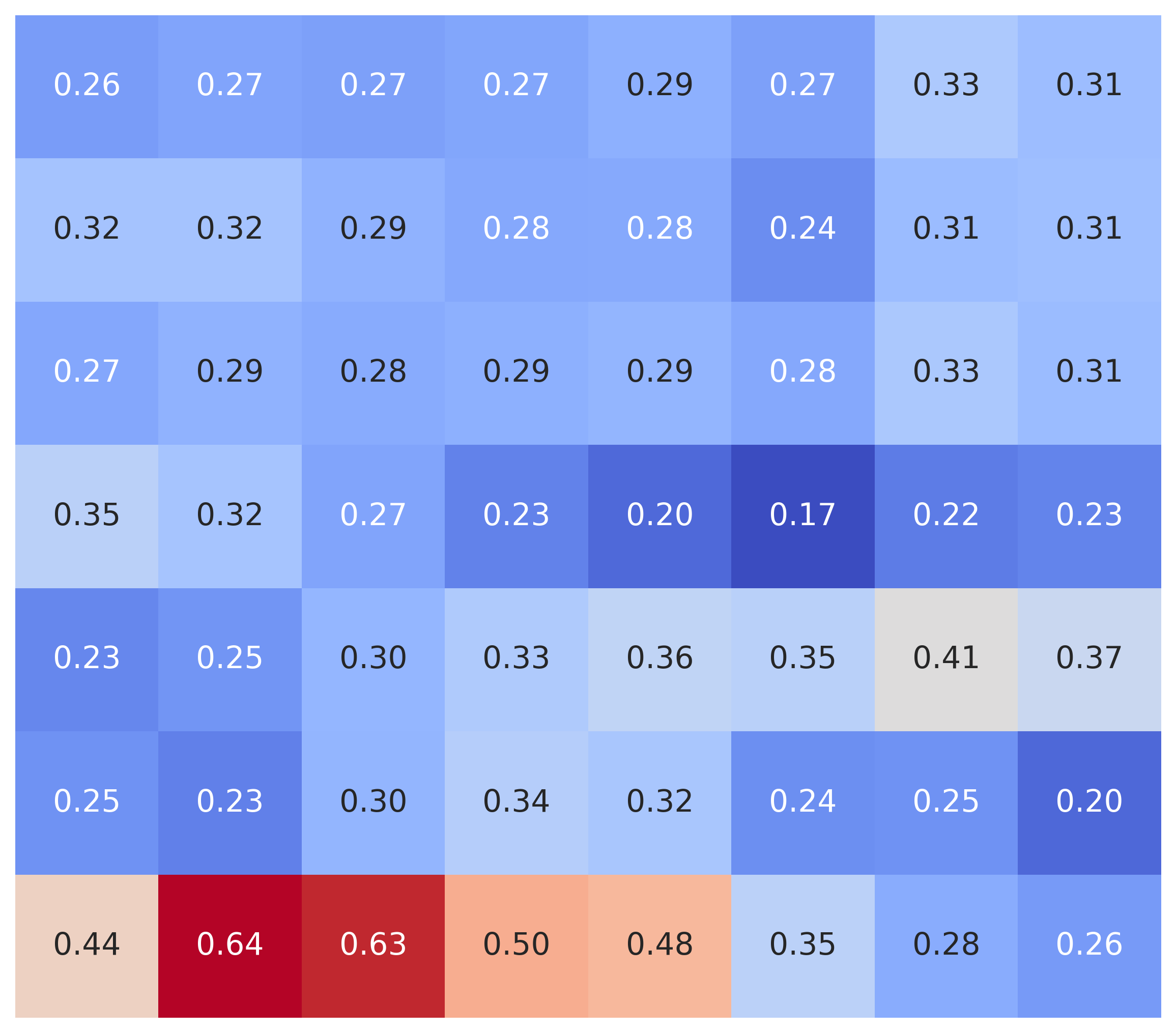}
        \vspace{-5mm}
        \caption{\scriptsize 336 Predicted Length}
    \end{subfigure}
    \hfill
    \begin{subfigure}[t]{0.205\textwidth}
        \includegraphics[width=\linewidth]{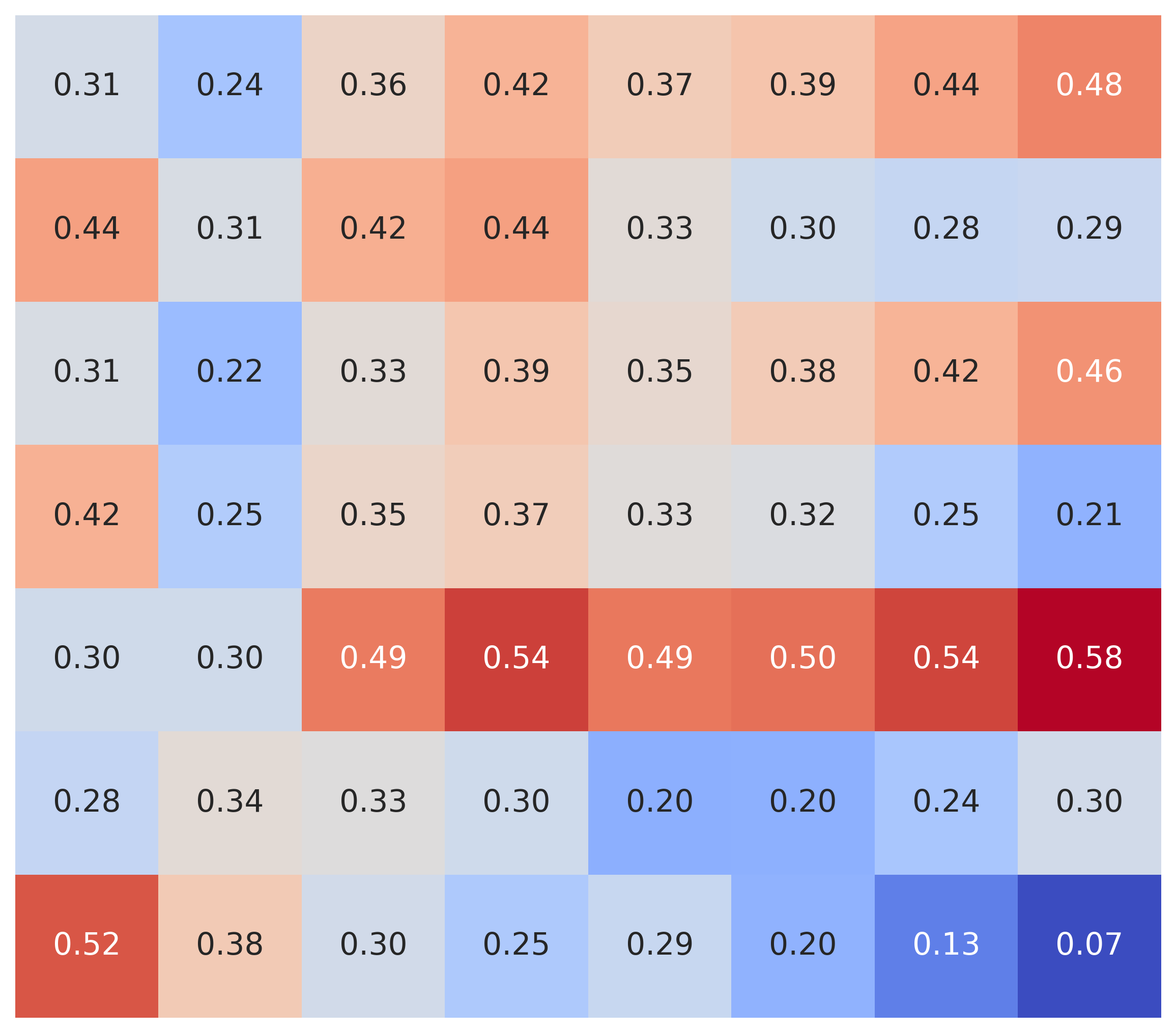}
        \vspace{-5mm}
        \caption{\scriptsize 720 Predicted Length}
    \end{subfigure}
    \caption{Heatmaps of STAM across different prediction lengths from the ETTh1. The horizontal axis is the value averaged over multiple prediction time points and the vertical axis refers to the different channels.}
    \label{fig:stam}
    \vspace{-4mm}
\end{figure}

\input{table/prompts}
\subsection{Model Analyses}

\subsubsection{Ablation Studies}

To ascertain the impact of different modules within DualSG, we perform ablation studies focusing on the following components. 

\noindent
(1) \textit{w/o TSGC:} Use only the numerical sequence stream by removing the dual stream and TSCG. Because it is not possible to use the SemFuse after remove the TSCG, CD is used.

\noindent
(2) \textit{w/o MAP:} Remove the Progressive Patch Tokenization Module.

\noindent
(3) \textit{w/o SemFuse:} Remove the Semantic Mediated Channel Fusion Module.

\noindent
(4) \textit{w/o LLM:} Remove the LLM Module.

Table \ref{tab:Ablationstudy} illustrates the unique impact of each module. We have the following observations: 1) Removing TSCG consistently degrades performance, with sharper drops on non-stationary datasets (e.g., Weather), confirming the value of semantic trend abstraction for stabilizing volatile patterns. 2) Replacing MAP with uniform patching harms accuracy, notably on high-frequency sequences (e.g., ETTm2), highlighting its strength in adaptive, multi-scale temporal modeling. 3) Substituting SemFuse with full attention or static clustering increases cost and reduces accuracy, especially on strongly correlated data, validating its efficiency in filtering spurious channel interactions via semantic guidance. 4) LLM is least useful because it only serves to reason and summarize, while the captions generated by TSCG already contains enough information. As shown in Table~\ref{tab:efficiency}, we evaluate on the ETTh2 dataset and find that caching pre-encoded time series significantly reduces training time without sacrificing predictive accuracy.

\subsubsection{Comparison of Prompt Strategies}
\label{sec:prompt_exp}
To explore the effectiveness of different textual inputs, we benchmark six prompt types under the same forecasting framework: domain, {timestamp}, statistics, instruction, external, and our proposed \texttt{caption}, as shown in Table \ref{tab:prompt}. 
As illustrated in Figure ~\ref{fig:prompt}, \texttt{caption}-based prompts consistently outperform others, achieving the lowest average MSE and the highest frequency of best results. The \texttt{timestamp} prompt ranks second, aligning with prior studies that highlight the benefit of temporal markers. Other prompt types suffer from lower semantic richness or contextual relevance. The \texttt{external} prompt category was excluded due to the absence of timestamp-aligned paired datasets.

\input{table/trend}

\subsubsection{Limitations of Fixed-length Patching and Different LLMs}
We conduct a comparative study based on PatchTS using four differ-
ent patch lengths: 16, 24, 36, and 48. Fixed patch lengths may overlook variable temporal patterns and dynamics. In addition, we assess the impact of different LLM backbones on TSC reasoning performance by integrating six representative open-source language models into the DualSG framework. These include {GPT2}, {BERT}, {DeepSeek-R1}, {Qwen2.5-3B}, {LLaMA2-7B}, and {LLaMA3.2-3B}, covering a range of architectures and parameter scales. This comparison provides insight into how different language models influence the semantic correction capability of the textual stream. We found in our experiment that GPT2 has the best effect.


\begin{table}[htbp]
\scriptsize
\centering
\caption{Comparison of training and inference efficiency.}
\label{tab:efficiency}
\begin{tabular}{lcccc}
\toprule
\textbf{Model} & \textbf{Train Time (min)} & \textbf{Inference Time (s)} & \textbf{MSE} & \textbf{MAE} \\
\midrule
DualSG           & 47.1 & 11.7 & 0.284 & 0.332 \\
DualSG (Cache)   & 13.3 & 11.7 & 0.284 & 0.332 \\
TimeCMA          & 59.0 &  4.8 & 0.329 & 0.365 \\
TimeLLM          & 95.0 & 90.8 & 0.317 & 0.370 \\
\bottomrule
\end{tabular}
\end{table}

\subsubsection{Comparsion Among Decomposed Components.}
A multivariate time series can be decomposed into three additive components: a trend component representing the overall directional movement, a seasonal component capturing periodic patterns, and a residual component modeling short-term noise or irregularities. To assess the comparative strength of DualSG in these components, we analyze the performance of the model as shown in Table \ref{tab:component_mae_ratio}. Due to the introduction of dual-stream collaboration, DualSG demonstrates the most pronounced improvement on the trend component, where accurate long-term pattern modeling is essential. Compared with baselines such as DLinear, PatchTST, and TimesNet, DualSG reduces trend-related MAE by up to 20.9\%. These findings support our claim that high-level semantic reasoning, introduced via time series captions, effectively complements numerical modeling in capturing trend dynamics. We attribute the decline in seasonal metrics in Table \ref{tab:component_mae_ratio} to the fixed lookback window size (96), which constrains the model’s capacity to capture recurring patterns. Although this issue diminishes under longer lookback configurations, we acknowledge that enhancing trend-level performance in DualSG may come at the expense of seasonal forecasting precision.

\input{table/full_long_exp}
\FloatBarrier

\subsubsection{Visualization of Spatial \& Temporal Attention Matrix.}

As shown in Figure \ref{fig:stam}, STAM exhibits increasingly distinct channel importance patterns as the prediction length grows. For short-term forecasts (e.g., 96 steps), the weights are relatively uniform across channels, indicating a balanced usage of input signals. In contrast, longer prediction horizons (e.g., 336 and 720 steps) reveal sharp attention concentration on specific channels, suggesting that STAM selectively relies on more informative variables when facing more challenging forecasting tasks. Also the longer the prediction length, the more weight is given to the textual information.

\subsubsection{Limitation}
While MAP alleviates part of the complexity, the dual-stream design still adds some overhead. In addition, the autoregressive nature of TSCG may affect inference speed. To maintain efficiency, we employ a lightweight pre-trained LLM.
\section{Conclusions}
In this paper, we address the key challenges of applying Large Language Models (LLMs) to multivariate time series forecasting, namely, two key limitations: numerical precision and alignment difficulty. We propose \textbf{DualSG}, a dual-stream forecasting framework that decouples numerical modeling from semantic reasoning. By introducing Time Series Caption as explicit semantic prompts and guiding multivariate interaction through caption-based fusion, DualSG achieves robust, interpretable, and accurate predictions. This work confirms that explicit semantic guidance, when properly integrated, can bridge the gap between LLMs and numerical forecasting tasks. In future work, we plan to extend this paradigm to more complex multimodal time series.


\begin{acks}
This work was supported by the Innovation Funding of Institute of Computing Technology, Chinese Academy of Sciences under Grant No. E461070.
\end{acks}

\bibliographystyle{ACM-Reference-Format}
\setcitestyle{numbers,sort&compress}
\balance
\bibliography{main}


\end{document}

%% file: table/ablation.tex
\renewcommand{\arraystretch}{1}
\begin{table}[t!]
\caption{Ablation study. Results are averaged from all forecasting horizons.}
\resizebox{1\columnwidth}{!}{
\footnotesize
\begin{tabular}{cc|cccccccccc}
\toprule
  \multicolumn{2}{c|}{\textbf{Model}} & \multicolumn{2}{c}{\textbf{DualSG}} & \multicolumn{2}{c}{\textbf{w/o  TSCG}} & \multicolumn{2}{c}{\textbf{w/o  MAP}} & \multicolumn{2}{c}{\textbf{w/o  SemFuse}} & \multicolumn{2}{c}{\textbf{w/o  LLM}}  \\
\multicolumn{2}{c|}{Metrics}  & mse & mae & mse & mae & mse & mae & mse & mae & mse & mae   \\
\midrule
\multicolumn{2}{c|}{ETTh1}&  \textbf{0.414} & \textbf{0.422} & 0.436 & 0.436 & 0.441 & 0.445 & 0.431 & 0.438 & 0.426 & 0.436 \\
\addlinespace\cline{1-12} \addlinespace
\multicolumn{2}{c|}{ETTh2}& \textbf{0.354} & \textbf{0.386} & 0.371 & 0.396 & 0.363 & 0.393 & 0.357 & 0.388 & 0.356 & 0.392 \\
\addlinespace\cline{1-12} \addlinespace
\multicolumn{2}{c|}{ETTm1}& \textbf{0.347} & \textbf{0.373} & 0.361 & 0.385 & 0.370 & 0.393 & 0.358 & 0.382 & 0.350 & 0.377 \\
\addlinespace\cline{1-12} \addlinespace 
\multicolumn{2}{c|}{ETTm2}& \textbf{0.254} & \textbf{0.308} & 0.264 & 0.318 & 0.279 & 0.326 & 0.258 & 0.312 & 0.256 & 0.315 \\
\addlinespace\cline{1-12} \addlinespace 
\multicolumn{2}{c|}{Weather}& \textbf{0.228} & \textbf{0.260} & 0.238 & 0.268 & 0.246 & 0.276 & 0.232 & 0.263 & 0.229 & 0.261 \\
\addlinespace
\bottomrule
\end{tabular}
}
\label{tab:Ablationstudy}
\end{table}

%% file: table/prompts.tex
\begin{table}[t!]
    \scriptsize
    \centering
    \caption{Six Types of Prompts for Time Series Forecasting.}
    \label{tab:prompt}
    \rowcolors{2}{gray!10}{white}
    \begin{tabular}{p{0.45\columnwidth} p{0.45\columnwidth}}
        \toprule
        \textbf{Type} & \textbf{Prompt Example} \\
        \midrule
        Domain & Weather is recorded every 10 minutes for the 2020 whole year, which contains 21 meteorological indicators, such as air temperature, humidity, etc. \\
        \midrule
        Timestamp & From August 16, 2019, Friday to August 30, 2019, Friday, the average temperature of region 110 was 78, 81, 83, 84, 84, 82, 83, 78, 77, 77, 74, 77, 78, 73, 76 degree on each day. \\
        \midrule
        Statistics & Input statistics: a) Min value: \(x_{\text{min}}\) ; b) Max value: \(x_{\text{max}}\); c) Median value: \(x_{\text{median}}\); d) Trend: upward or downward; e) Top 5 lags: \(x_{\text{top}}\) \\
        \midrule
        Instruction & Predict the future time step given the trend, season and residual. \\
        \midrule
        External & The National Guard provides support to state emergency management operations under the control of the Governor and state Adjutant General.\\
        \midrule
        Caption & The time series exhibits a gradual decrease from start to end, with a flat end. \\
        \bottomrule
    \end{tabular}
\end{table}


%% file: table/trend.tex
\begin{table}[ht]
\centering
\footnotesize
\caption{Average MAE and relative improvement (\% worse than DualSG) on decomposed components. DualSG achieves the most significant gain on trend modeling. Here, DualSG's lookback windows $L=96$, so it doesn't perform optimally on all components. The Weather dataset is used for this experiment.}
\label{tab:component_mae_ratio}
\begin{tabular}{lccc}
\toprule
\textbf{Method} & \textbf{Trend MAE} (\textdownarrow) & \textbf{Seasonal MAE} (\textdownarrow) & \textbf{Residual MAE} (\textdownarrow) \\
\midrule
DualSG (Ours)      & 0.239  & 0.0113   & 0.0388   \\
\rowcolor{gray!10}
DLinear         & 0.289 \textcolor{gray}{(+20.9\%)} & 0.0105 \textcolor{gray}{(-7.1\%)} & 0.0335 \textcolor{gray}{(-13.7\%)} \\
\rowcolor{gray!10}
PatchTST        & 0.272 \textcolor{gray}{(+13.8\%)} & 0.0113 \textcolor{gray}{(0.0\%)} & 0.0388 \textcolor{gray}{(0.0\%)} \\
\rowcolor{gray!10}
TimesNet        & 0.263 \textcolor{gray}{(+10.0\%)} & 0.0110 \textcolor{gray}{(-2.7\%)} & 0.0380 \textcolor{gray}{(-2.1\%)} \\
\bottomrule
\end{tabular}
\end{table}

%% file: table/full_long_exp.tex
\begin{table*}[!ht] 
	\setlength{\tabcolsep}{1.35pt}
	\scriptsize
	\centering
    \caption{Full results for long-term forecasting consider prediction horizons $H$ within $\{96, 192, 336, 512\}$. The lookback window for our framework is selected via a search over $\{96, 192, 336, 512, 720\}$. The best and second best outcomes are highlighted in \best{best} and \second{second}, respectively. The notation "$1^{\text{st}}$ \textit{Count}" denotes the frequency of each method achieving the top results.}
	\begin{threeparttable}
		\begin{tabular}{c|c|c c|c c|c c|c c|c c|c c|c c|c c|c c|c c|c c|c c|c c|c c}
			\toprule

            \multicolumn{2}{c}{Categories} & \multicolumn{8}{c}{LLM-based} & \multicolumn{12}{c}{Transformer-based} & \multicolumn{4}{c}{CNN-based} & \multicolumn{4}{c}{MLP-based} \\
            
			\toprule
			\multicolumn{2}{c}{\multirow{2}{*}{\scalebox{1.1}{Models}}}& \multicolumn{2}{c}{DualSG} & \multicolumn{2}{c}{CALF} & \multicolumn{2}{c}{TimeLLM} & \multicolumn{2}{c}{GPT4TS} & \multicolumn{2}{c}{DUET} & \multicolumn{2}{c}{PatchTST} & \multicolumn{2}{c}{iTransformer} & \multicolumn{2}{c}{Crossformer} & \multicolumn{2}{c}{FEDformer} & \multicolumn{2}{c}{Autoformer} & \multicolumn{2}{c}{TimesNet} & \multicolumn{2}{c}{MICN} & \multicolumn{2}{c}{DLinear} & \multicolumn{2}{c}{TimeMixer} \\
			\multicolumn{2}{c}{} & \multicolumn{2}{c}{\scalebox{0.8}{\textbf{Ours}}} & \multicolumn{2}{c}{\scalebox{0.8}{\citeyearpar{calf}}} & \multicolumn{2}{c}{\scalebox{0.8}{\citeyearpar{timellm}}} & \multicolumn{2}{c}{\scalebox{0.8}{\citeyearpar{gpt4ts}}} & \multicolumn{2}{c}{\scalebox{0.8}{\citeyearpar{qiu2025duet}}} & \multicolumn{2}{c}{\scalebox{0.8}{\citeyearpar{patchtst}}} & \multicolumn{2}{c}{\scalebox{0.8}{\citeyearpar{liu2024itransformer}}} & \multicolumn{2}{c}{\scalebox{0.8}{\citeyearpar{zhang2023crossformer}}} & \multicolumn{2}{c}{\scalebox{0.8}{\citeyearpar{zhou2022fedformer}}} & \multicolumn{2}{c}{\scalebox{0.8}{\citeyearpar{wu2021autoformer}}} & \multicolumn{2}{c}{\scalebox{0.8}{\citeyearpar{timesnet}}} & \multicolumn{2}{c}{\scalebox{0.8}{\citeyearpar{wang2023micn}}} & \multicolumn{2}{c}{\scalebox{0.8}{\citeyearpar{dlinear}}} & \multicolumn{2}{c}{\scalebox{0.8}{\citeyearpar{wang2024timemixer}}} \\
			\cmidrule(lr){3-4} \cmidrule(lr){5-6} \cmidrule(lr){7-8} \cmidrule(lr){9-10} \cmidrule(lr){11-12} \cmidrule(lr){13-14} \cmidrule(lr){15-16} \cmidrule(lr){17-18} \cmidrule(lr){19-20} \cmidrule(lr){21-22} \cmidrule(lr){23-24} \cmidrule(lr){25-26} \cmidrule(lr){27-28} \cmidrule(lr){29-30} 
			\multicolumn{2}{c}{Metric}& MSE & MAE & MSE & MAE & MSE & MAE & MSE & MAE & MSE & MAE & MSE & MAE & MSE & MAE & MSE & MAE & MSE & MAE & MSE & MAE & MSE & MAE & MSE & MAE & MSE & MAE & MSE & MAE \\
			\toprule
			\multirow{5}{*}{\rotatebox[origin=c]{90}{ETTm1}} 
			& 96 & \best{0.285} & \best{0.330} & 0.323 & 0.349 & 0.359 & 0.381 & 0.329 & 0.364 & \second{0.295} & \second{0.336} & 0.321 & 0.360 & 0.341 & 0.376 & 0.360 & 0.401 & 0.379 & 0.419 & 0.505 & 0.475 & 0.338 & 0.375 & 0.316 & 0.362 & 0.345 & 0.372 & 0.320 & 0.357 \\
			& 192 & \best{0.319} & \best{0.356} & 0.374 & 0.375 & 0.383 & 0.393 & 0.368 & 0.382 & \second{0.338} & \second{0.366} & 0.362 & 0.384  & 0.382 & 0.395 & 0.402 & 0.440 & 0.426 & 0.441 & 0.553 & 0.496 & 0.374 & 0.387 & 0.363 & 0.390 & 0.380 & 0.389 & 0.361 & 0.381 \\
			& 336 & \best{0.366} & \best{0.388}  & 0.409 & 0.399 & 0.416 & 0.414 & 0.400 & 0.403 & \second{0.364} & \second{0.386} & 0.392 & 0.402 & 0.418 & 0.418 & 0.543 & 0.528 & 0.445 & 0.459 & 0.621 & 0.537 & 0.410 & 0.411 & 0.408 & 0.426 & 0.413 & 0.413 & 0.390 & 0.404 \\
			& 720 & \best{0.417} & \best{0.417}  & 0.477 & 0.438 & 0.483 & 0.449 & 0.460 & 0.439 & \second{0.411} & \second{0.415} & 0.450 & 0.435 & 0.487 & 0.456 & 0.704 & 0.642 & 0.543 & 0.490 & 0.671 & 0.561 & 0.478 & 0.450 & 0.481 & 0.476 & 0.474 & 0.453 & 0.454 & 0.441 \\
			\cmidrule(lr){2-30}
			& \emph{Avg.} &\best{0.347} &\best{0.372}& 0.395 & 0.390 & 0.410 & 0.409 & 0.389 & 0.397 & \second{0.352} & \second{0.376} & 0.381 & 0.395 & 0.407 & 0.411 & 0.502 & 0.502 & 0.448 & 0.452 & 0.588 & 0.517 & 0.400 & 0.406 & 0.392 & 0.413 & 0.403 & 0.407 & 0.381 & 0.395 \\
			\midrule
			\multirow{5}{*}{\rotatebox[origin=c]{90}{ETTm2}} 
			& 96 &\best{0.165} &\best{0.247}  & 0.178 & 0.256 & 0.193 & 0.280 & 0.178 & 0.263 & \second{0.166} & \second{0.253} & 0.178 & 0.260 & 0.185 & 0.272 & 0.273 & 0.356 & 0.203 & 0.287 & 0.255 & 0.339 & 0.187 & 0.267 & 0.179 & 0.275 & 0.193 & 0.292 & 0.175 & 0.258 \\
			& 192 &\best{0.219} &\best{0.286} & 0.242 & \second{0.297} & 0.257 & 0.318 & 0.245 & 0.306 & 0.240 & 0.300 & 0.249 & 0.307 & 0.253 & 0.313 & 0.426 & 0.487 & 0.269 & 0.328 & 0.249 & 0.309 & 0.533 & 0.563 & 0.307 & 0.376 & 0.284 & 0.362 & \second{0.237} & 0.299 \\
			& 336 &\best{0.271} &\best{0.320} & 0.307 & 0.339 & 0.317 & 0.353 & 0.309 & 0.347 & \second{0.294} & \second{0.335} & 0.313 & 0.346 & 0.315 & 0.350 & 1.013 & 0.714 & 0.325 & 0.366 & 0.339 & 0.372 & 0.321 & 0.351 & 0.325 & 0.388 & 0.369 & 0.427 & 0.298 & 0.340 \\
			& 720 & \best{0.363} & \best{0.380}  & 0.397 & 0.393 & 0.419 & 0.411 & 0.409 & 0.408 & \second{0.371} & \second{0.383} & 0.400 & 0.398 & 0.413 & 0.406 & 3.154 & 1.274 & 0.421 & 0.415 & 0.433 & 0.432 & 0.408 & 0.403 & 0.502 & 0.490 & 0.554 & 0.522 & 0.391 & 0.396 \\
			\cmidrule(lr){2-30}
			& \emph{Avg.}&\best{0.254}&\best{0.308} & 0.281 & 0.321 & 0.296 & 0.340 & 0.285 & 0.331 & \second{0.267} & \second{0.318} & 0.285 & 0.327 & 0.291 & 0.335 & 1.216 & 0.707 & 0.305 & 0.349 & 0.327 & 0.371 & 0.291 & 0.333 & 0.328 & 0.382 & 0.350 & 0.401 & 0.275 & 0.323 \\
			\midrule
			\multirow{5}{*}{\rotatebox[origin=c]{90}{ETTh1}} 
			& 96 &\second{0.369} &\best{0.386}  & \second{0.369} & \second{0.389} & 0.398 & 0.410 & 0.376 & 0.397 & \best{0.360} & \second{0.389} & 0.393 & 0.408 & 0.386 & 0.404 & 0.420 & 0.439 & 0.376 & 0.419 & 0.449 & 0.459 & 0.384 & 0.402 & 0.421 & 0.431 & 0.386 & 0.400 & 0.375 & 0.400 \\
			& 192 &\best{0.396} &\best{0.417} & 0.427 & 0.423 & 0.451 & 0.440 & 0.438 & 0.426 & \second{0.407} & 0.425 & 0.445 & 0.434 & 0.441 & 0.436 & 0.540 & 0.519 & 0.420 & 0.448 & 0.436 & 0.429 & 1.008 & 0.792 & 0.474 & 0.487 & 0.437 & 0.432 & 0.429 & \second{0.421} \\
			& 336 &\best{0.420} &\best{0.426} & 0.456 & \second{0.436} & 0.508 & 0.471 & 0.479 & 0.446 & \second{0.430} & 0.443 & 0.484 & 0.451 & 0.489 & 0.461 & 0.722 & 0.648 & 0.459 & 0.465 & 0.521 & 0.496 & 0.491 & 0.469 & 0.569 & 0.551 & 0.481 & 0.459 & 0.484 & 0.458 \\
			& 720 &\best{0.470} &\best{0.460} & 0.479 & \second{0.467} & 0.483 & 0.478 & 0.495 & 0.476 & \second{0.450} & 0.471 & 0.480 & 0.471 & 0.508 & 0.493 & 0.799 & 0.685 & 0.506 & 0.507 & 0.514 & 0.512 & 0.521 & 0.500 & 0.770 & 0.672 & 0.519 & 0.516 & 0.498 & 0.482 \\
			\cmidrule(lr){2-30}
			& \emph{Avg.} &\best{0.413} &\best{0.422}& 0.432 & \second{0.428} & 0.460 & 0.449 & 0.447 & 0.436 & \second{0.412} & 0.432 & 0.450 & 0.441 & 0.455 & 0.448 & 0.620 & 0.572 & 0.440 & 0.460 & 0.496 & 0.487 & 0.458 & 0.450 & 0.558 & 0.535 & 0.456 & 0.452 & 0.447 & 0.440 \\
			\midrule
			\multirow{5}{*}{\rotatebox[origin=c]{90}{ETTh2}} 
			& 96 &\second{0.284} &\second{0.332}  & \best{0.279} & \best{0.331} & 0.295 & 0.346 & 0.295 & 0.348 & 0.337 & 0.367 & 0.294 & 0.343  & 0.300 & 0.349 & 0.745 & 0.584 & 0.358 & 0.397 & 0.346 & 0.388 & 0.340 & 0.374 & 0.299 & 0.364 & 0.333 & 0.387 & 0.289 & 0.341 \\
			& 192 &\best{0.348} &\best{0.378} & \second{0.353} & \second{0.380} & 0.386 & 0.399 & 0.386 & 0.404 & 0.381 & 0.402 & 0.377 & 0.393 & 0.379 & 0.398 & 0.877 & 0.656 & 0.429 & 0.439 & 0.456 & 0.452 & 0.402 & 0.414 & 0.441 & 0.454 & 0.477 & 0.476 & 0.372 & 0.392 \\
			& 336 &\second{0.370} &\second{0.400} & \best{0.362} & \best{0.394} & 0.447 & 0.443 & 0.421 & 0.435 & 0.400 & 0.420 & 0.381 & 0.409 & 0.418 & 0.429 & 1.043 & 0.731 & 0.496 & 0.487 & 0.482 & 0.486 & 0.452 & 0.452 & 0.654 & 0.567 & 0.594 & 0.541 & 0.386 & 0.414 \\
			& 720&0.412 &\second{0.433}  & \best{0.404} & \best{0.426} & 0.428 & 0.444 & 0.422 & 0.445 & 0.443 & 0.457 & \second{0.412} & \second{0.433} & 0.428 & 0.445 & 1.104 & 0.763 & 0.463 & 0.474 & 0.515 & 0.511 & 0.462 & 0.468 & 0.956 & 0.716 & 0.831 & 0.657 & 0.412 & 0.434 \\
			\cmidrule(lr){2-30}
			& \emph{Avg.}&\second{0.353} &\second{0.389} & \best{0.349} & \best{0.382} & 0.389 & 0.408 & 0.381 & 0.408 & 0.390 & 0.412 & 0.366 & 0.394 & 0.381 & 0.405 & 0.942 & 0.684 & 0.437 & 0.449 & 0.450 & 0.459 & 0.414 & 0.427 & 0.587 & 0.525 & 0.559 & 0.515 & 0.364 & 0.395 \\
			\midrule
			\multirow{5}{*}{\rotatebox[origin=c]{90}{Weather}}
            
			& 96  &\second{0.156} & \second{0.201} & 0.164 & 0.204 & 0.195 & 0.233 & 0.182 & 0.223 & \best{0.152} & \best{0.196} & 0.177 & 0.218 & 0.174 & 0.214 &  0.158 & 0.230 & 0.217 & 0.296 & 0.266 & 0.336 & 0.172 & 0.220 & 0.161 & 0.229 & 0.196 & 0.255 & 0.163 & 0.209 \\
            
			& 192 &\best{0.196} &\best{0.239} & 0.214 & 0.250 & 0.240 & 0.269 & 0.231 & 0.263 & \second{0.198} & \second{0.240} & 0.225 & 0.259 & 0.221 & 0.254 & \second{0.206} & 0.277 & 0.276 & 0.336 & 0.307 & 0.367 & 0.219 & 0.261 & 0.220 & 0.281 & 0.237 & 0.296 & 0.208 & 0.250 \\
            
			& 336&\best{0.248} &\best{0.280}  & 0.269 & 0.291 & 0.293 & 0.306 & 0.283 & 0.300 & \second{0.249} & \best{0.280} & 0.278 & 0.297 & 0.278 & 0.296 & 0.272 & 0.335 & 0.339 & 0.380 & 0.359 & 0.395 & 0.280 & 0.306 & 0.278 & 0.331 & 0.283 & 0.335 & 0.251 & \second{0.287} \\
            
			& 720 &\second{0.323} &\best{0.334} & 0.355 & 0.352 & 0.368 & 0.354 & 0.360 & 0.350 & 0.332 & \second{0.336} & 0.354 & 0.348 & 0.358 & 0.349 & 0.398 & 0.418 & 0.403 & 0.428 & 0.419 & 0.428 & 0.365 & 0.359 & \best{0.311} & 0.356 & 0.345 & 0.381 & 0.339 & \second{0.341} \\
			\cmidrule(lr){2-30}
            
			& \emph{Avg.}&\best{0.230} &\best{0.263} & 0.250 & 0.274 & 0.274 & 0.290 & 0.264 & 0.284 & \second{0.233} & \best{0.263} & 0.258 & 0.280 & 0.257 & 0.279 & 0.259 & 0.315 & 0.309 & 0.360 & 0.338 & 0.382 & 0.259 & 0.287 & 0.242 & 0.299 & 0.265 & 0.317 & 0.240 & \second{0.271} \\
			\midrule
			\multirow{5}{*}{\rotatebox[origin=c]{90}{Electricity}}
			& 96  &\best{0.131} &\best{0.222}  & 0.145 & 0.238 & 0.204 & 0.293 & 0.185 & 0.272 & \second{0.134} & \second{0.234} & 0.195 & 0.285 & 0.148 & 0.240 & 0.219 & 0.314 & 0.193 & 0.308 & 0.201 & 0.317 & 0.168 & 0.272 & 0.164 & 0.269 & 0.197 & 0.282 & 0.153 & 0.247 \\
			& 192 &\best{0.151} &\best{0.240} & 0.161 & \second{0.252} & 0.207 & 0.295 & 0.189 & 0.276 & \second{0.153} & \second{0.252} & 0.199 & 0.289 & 0.162 & 0.253 & 0.231 & 0.322 & 0.201 & 0.315 & 0.222 & 0.334 & 0.184 & 0.289 & 0.177 & 0.285 & 0.196 & 0.285 & 0.166 & 0.256 \\
			& 336 &\best{0.162} &\best{0.256} & 0.175 & \second{0.267} & 0.219 & 0.308 & 0.204 & 0.291 & \second{0.170} & 0.268 & 0.215 & 0.305 & 0.178 & 0.269 & 0.246 & 0.337 & 0.214 & 0.329 & 0.231 & 0.338 & 0.198 & 0.300 & 0.193 & 0.304 & 0.209 & 0.301 & 0.185 & 0.277 \\
			& 720 &\second{0.208} &\best{0.296}   & 0.222 & \second{0.303} & 0.263 & 0.341 & 0.245 & 0.324 & \best{0.204} & \best{0.296} & 0.256 & 0.337 & 0.225 & 0.317 & 0.280 & 0.363 & 0.246 & 0.355 & 0.254 & 0.361 & 0.220 & 0.320 & 0.212 & 0.321 & 0.245 & 0.333 & 0.225 & 0.310 \\
			\cmidrule(lr){2-30}
			& \emph{Avg.} &\best{0.163} &\best{0.254} & 0.175 & 0.265 & 0.223 & 0.309 & 0.205 & 0.290 & \second{0.165} & \second{0.263} & 0.216 & 0.304 & 0.178 & 0.270 & 0.244 & 0.334 & 0.214 & 0.327 & 0.227 & 0.338 & 0.192 & 0.295 & 0.186 & 0.294 & 0.212 & 0.300 & 0.216 & 0.280 \\
			\midrule
			\multirow{5}{*}{\rotatebox[origin=c]{90}{Traffic}}
			& 96&\best{0.365} &\best{0.254}   & 0.407 & 0.268 & 0.536 & 0.359 & 0.468 & 0.307 & \second{0.371} & \second{0.255} & 0.544 & 0.359 & 0.395 & 0.268 &  0.522 & 0.290 & 0.587 & 0.366 & 0.613 & 0.388 & 0.593 & 0.321 & 0.519 & 0.309 & 0.650 & 0.396 & 0.462 & 0.285 \\
			& 192 & \best{0.385} &\second{0.268} & 0.430 & 0.278 & 0.530 & 0.354 & 0.476 & 0.311 & \second{0.395} & \best{0.265} & 0.540 & 0.354 & 0.417 & 0.276 & 0.530 & 0.293 & 0.604 & 0.373 & 0.616 & 0.382 & 0.617 & 0.336 & 0.537 & 0.315 & 0.598 & 0.370 & 0.473 & 0.296 \\
			& 336 & \best{0.394} & \best{0.275}  & 0.444 & 0.281 & 0.530 & 0.349 & 0.488 & 0.317 & \second{0.413} & \second{0.276} & 0.551 & 0.358 & 0.433 & 0.283 & 0.558 & 0.305 & 0.621 & 0.383 & 0.622 & 0.337 & 0.629 & 0.336 & 0.534 & 0.313 & 0.605 & 0.373 & 0.498 & 0.296 \\
			& 720&\best{0.439} &\second{0.299}  & 0.477 & 0.300 & 0.569 & 0.371 & 0.521 & 0.333 & \second{0.457} & \best{0.290} & 0.586 & 0.375 & 0.467 & 0.302 & 0.589 & 0.328 & 0.626 & 0.382 & 0.660 & 0.408 & 0.640 & 0.350 & 0.577 & 0.325 & 0.645 & 0.394 & 0.506 & 0.313 \\
			\cmidrule(lr){2-30}
			& \emph{Avg.}&\best{0.396} &\best{0.274} & 0.439 & 0.281 & 0.541 & 0.358 & 0.488 & 0.317 & \second{0.409} & \second{0.276} & 0.555 & 0.361 & 0.428 & 0.282 & 0.550 & 0.304 & 0.610 & 0.376 & 0.628 & 0.379 & 0.620 & 0.336 & 0.541 & 0.315 & 0.625 & 0.383 & 0.484 & 0.297 \\
            \midrule
            \multirow{5}{*}{\rotatebox[origin=c]{90}{Solar-Energy}}
			& 96 & \best{0.169} &\best{0.223}  & 0.283 & 0.353 & 0.293 & 0.381 & 0.268 & 0.357 & 0.193 & \best{0.223} & \second{0.170} & \second{0.234} & 0.203 & 0.237 &  0.310 & 0.331 & 0.242 & 0.342 & 0.252 & 0.351 & 0.250 & 0.292 & 0.263 & 0.307 & 0.290 & 0.378 & 0.189 & 0.259 \\
			& 192 & \best{0.185} &\best{0.241} & 0.316 & 0.374 & 0.329 & 0.384 & 0.276 & 0.331 & 0.238 & 0.287 & \second{0.204} & 0.302 & 0.233 & \second{0.261} & 0.734 & 0.725  & 0.285 & 0.380 & 0.289 & 0.384 & 0.296 & 0.318 & 0.314 & 0.328 & 0.320 & 0.398 & 0.222 & 0.283 \\
			& 336 & \best{0.200} &\best{0.265}  & 0.347 & 0.393 & 0.366 & 0.427 & 0.288 & 0.367 & 0.235 & 0.282 & \second{0.212} & 0.293 & 0.248 & \second{0.273} & 0.750 & 0.735  & 0.282 & 0.326 & 0.288 & 0.385 & 0.319 & 0.330 & 0.324 & 0.343 & 0.353 & 0.415 & 0.231 & 0.292 \\
			& 720 & \best{0.215} &\best{0.293}  & 0.348 & 0.389 & 0.371 & 0.415 & 0.321 & 0.363 & 0.236 & 0.290 & 0.220 & 0.317 & 0.250 & \second{0.281} & 0.769 & 0.765  & 0.357 & 0.427 & 0.366 & 0.436 & 0.338 & 0.337 & 0.349 & 0.345 & 0.357 & 0.413 & \second{0.223} & 0.285 \\
			\cmidrule(lr){2-30}
			& \emph{Avg.} &\best{0.192} &\best{0.259} & 0.324 & 0.377 & 0.340 & 0.401 & 0.288 & 0.355 & \second{0.228} & 0.280 & \second{0.200} & 0.284 & 0.233 & \second{0.262} & 0.641 & 0.639  & 0.292 & 0.381 & 0.298 & 0.386 & 0.301 & 0.319 & 0.314 & 0.332 & 0.330 & 0.401 & 0.216 & 0.280 \\
            \midrule
            \multicolumn{2}{c}{$1^{\text{st}}$ \emph{Count}} & \multicolumn{2}{c}{\best{65}}& \multicolumn{2}{c}{8} & \multicolumn{2}{c}{0} & \multicolumn{2}{c}{0} & \multicolumn{2}{c}{\second{10}} & \multicolumn{2}{c}{0} & \multicolumn{2}{c}{0} & \multicolumn{2}{c}{1} & \multicolumn{2}{c}{0} & \multicolumn{2}{c}{0} & \multicolumn{2}{c}{0} & \multicolumn{2}{c}{1} & \multicolumn{2}{c}{0} & \multicolumn{2}{c}{0} \\
			\toprule
		\end{tabular}
	\end{threeparttable}
	\label{tab::long-term}
\end{table*}

%% file: sample-sigconf.bbl

\begin{thebibliography}{64}


\ifx \showCODEN    \undefined \def \showCODEN     #1{\unskip}     \fi
\ifx \showDOI      \undefined \def \showDOI       #1{#1}\fi
\ifx \showISBNx    \undefined \def \showISBNx     #1{\unskip}     \fi
\ifx \showISBNxiii \undefined \def \showISBNxiii  #1{\unskip}     \fi
\ifx \showISSN     \undefined \def \showISSN      #1{\unskip}     \fi
\ifx \showLCCN     \undefined \def \showLCCN      #1{\unskip}     \fi
\ifx \shownote     \undefined \def \shownote      #1{#1}          \fi
\ifx \showarticletitle \undefined \def \showarticletitle #1{#1}   \fi
\ifx \showURL      \undefined \def \showURL       {\relax}        \fi
\providecommand\bibfield[2]{#2}
\providecommand\bibinfo[2]{#2}
\providecommand\natexlab[1]{#1}
\providecommand\showeprint[2][]{arXiv:#2}

\bibitem[Bao et~al\mbox{.}(2024)]%
        {bao2024boundaryaware}
\bibfield{author}{\bibinfo{person}{Yiying Bao}, \bibinfo{person}{Hao Zhou}, \bibinfo{person}{Chao Peng}, \bibinfo{person}{Chenyang Xu}, \bibinfo{person}{Shuo Shi}, {and} \bibinfo{person}{Kecheng Cai}.} \bibinfo{year}{2024}\natexlab{}.
\newblock \showarticletitle{Boundary-Aware Periodicity-based Sparsification Strategy for Ultra-Long Time Series Forecasting}. In \bibinfo{booktitle}{\emph{ACM Multimedia 2024}}.
\newblock


\bibitem[Bi et~al\mbox{.}(2023)]%
        {bi2023}
\bibfield{author}{\bibinfo{person}{Kaifeng Bi}, \bibinfo{person}{Lingxi Xie}, \bibinfo{person}{Hengheng Zhang}, \bibinfo{person}{Xin Chen}, \bibinfo{person}{Xiaotao Gu}, {and} \bibinfo{person}{Qi Tian}.} \bibinfo{year}{2023}\natexlab{}.
\newblock \showarticletitle{Accurate medium-range global weather forecasting with 3D neural networks}.
\newblock \bibinfo{journal}{\emph{Nature}} \bibinfo{volume}{619}, \bibinfo{number}{7970} (\bibinfo{date}{01 07} \bibinfo{year}{2023}), \bibinfo{pages}{533--538}.
\newblock
\showISSN{1476-4687}
\urldef\tempurl%
\url{https://doi.org/10.1038/s41586-023-06185-3}
\showDOI{\tempurl}


\bibitem[Bostrom and Durrett(2020)]%
        {bostrom2020bytepairencodingsuboptimal}
\bibfield{author}{\bibinfo{person}{Kaj Bostrom} {and} \bibinfo{person}{Greg Durrett}.} \bibinfo{year}{2020}\natexlab{}.
\newblock \bibinfo{title}{Byte Pair Encoding is Suboptimal for Language Model Pretraining}.
\newblock
\newblock
\showeprint[arxiv]{2004.03720}~[cs.CL]


\bibitem[Box and Pierce(1970)]%
        {box1970distribution}
\bibfield{author}{\bibinfo{person}{George~EP Box} {and} \bibinfo{person}{David~A Pierce}.} \bibinfo{year}{1970}\natexlab{}.
\newblock \showarticletitle{Distribution of residual autocorrelations in autoregressive-integrated moving average time series models}.
\newblock \bibinfo{journal}{\emph{Journal of the American statistical Association}} \bibinfo{volume}{65}, \bibinfo{number}{332} (\bibinfo{year}{1970}), \bibinfo{pages}{1509--1526}.
\newblock


\bibitem[Chen et~al\mbox{.}(2024)]%
        {chen2024pathformer}
\bibfield{author}{\bibinfo{person}{Peng Chen}, \bibinfo{person}{Yingying ZHANG}, \bibinfo{person}{Yunyao Cheng}, \bibinfo{person}{Yang Shu}, \bibinfo{person}{Yihang Wang}, \bibinfo{person}{Qingsong Wen}, \bibinfo{person}{Bin Yang}, {and} \bibinfo{person}{Chenjuan Guo}.} \bibinfo{year}{2024}\natexlab{}.
\newblock \showarticletitle{Pathformer: Multi-scale Transformers with Adaptive Pathways for Time Series Forecasting}. In \bibinfo{booktitle}{\emph{The Twelfth International Conference on Learning Representations}}.
\newblock


\bibitem[Cleveland et~al\mbox{.}(1990)]%
        {cleveland1990stl}
\bibfield{author}{\bibinfo{person}{Robert~B Cleveland}, \bibinfo{person}{William~S Cleveland}, \bibinfo{person}{Jean~E McRae}, \bibinfo{person}{Irma Terpenning}, {et~al\mbox{.}}} \bibinfo{year}{1990}\natexlab{}.
\newblock \showarticletitle{STL: A seasonal-trend decomposition}.
\newblock \bibinfo{journal}{\emph{J. off. Stat}} \bibinfo{volume}{6}, \bibinfo{number}{1} (\bibinfo{year}{1990}), \bibinfo{pages}{3--73}.
\newblock


\bibitem[Das et~al\mbox{.}(2023)]%
        {das2023longterm}
\bibfield{author}{\bibinfo{person}{Abhimanyu Das}, \bibinfo{person}{Weihao Kong}, \bibinfo{person}{Andrew Leach}, \bibinfo{person}{Shaan~K Mathur}, \bibinfo{person}{Rajat Sen}, {and} \bibinfo{person}{Rose Yu}.} \bibinfo{year}{2023}\natexlab{}.
\newblock \showarticletitle{Long-term Forecasting with Ti{DE}: Time-series Dense Encoder}.
\newblock \bibinfo{journal}{\emph{Transactions on Machine Learning Research}} (\bibinfo{year}{2023}).
\newblock
\showISSN{2835-8856}


\bibitem[Das et~al\mbox{.}(2024)]%
        {das2024decoderonlyfoundationmodeltimeseries}
\bibfield{author}{\bibinfo{person}{Abhimanyu Das}, \bibinfo{person}{Weihao Kong}, \bibinfo{person}{Rajat Sen}, {and} \bibinfo{person}{Yichen Zhou}.} \bibinfo{year}{2024}\natexlab{}.
\newblock \bibinfo{title}{A decoder-only foundation model for time-series forecasting}.
\newblock
\newblock
\showeprint[arxiv]{2310.10688}~[cs.CL]


\bibitem[Gruver et~al\mbox{.}(2023)]%
        {llmtime}
\bibfield{author}{\bibinfo{person}{Nate Gruver}, \bibinfo{person}{Marc Finzi}, \bibinfo{person}{Shikai Qiu}, {and} \bibinfo{person}{Andrew~Gordon Wilson}.} \bibinfo{year}{2023}\natexlab{}.
\newblock \showarticletitle{Large language models are zero-shot time series forecasters}. In \bibinfo{booktitle}{\emph{Proceedings of the 37th International Conference on Neural Information Processing Systems}} (New Orleans, LA, USA) \emph{(\bibinfo{series}{NIPS '23})}. \bibinfo{publisher}{Curran Associates Inc.}, \bibinfo{address}{Red Hook, NY, USA}, Article \bibinfo{articleno}{861}, \bibinfo{numpages}{14}~pages.
\newblock


\bibitem[Hu et~al\mbox{.}(2025)]%
        {hu2025contextalignment}
\bibfield{author}{\bibinfo{person}{Yuxiao Hu}, \bibinfo{person}{Qian Li}, \bibinfo{person}{Dongxiao Zhang}, \bibinfo{person}{Jinyue Yan}, {and} \bibinfo{person}{Yuntian Chen}.} \bibinfo{year}{2025}\natexlab{}.
\newblock \showarticletitle{Context-Alignment: Activating and Enhancing {LLM}s Capabilities in Time Series}. In \bibinfo{booktitle}{\emph{The Thirteenth International Conference on Learning Representations}}.
\newblock


\bibitem[Huang et~al\mbox{.}(2024)]%
        {DBLP:conf/aaai/HuangSZCDZW24}
\bibfield{author}{\bibinfo{person}{Qihe Huang}, \bibinfo{person}{Lei Shen}, \bibinfo{person}{Ruixin Zhang}, \bibinfo{person}{Jiahuan Cheng}, \bibinfo{person}{Shouhong Ding}, \bibinfo{person}{Zhengyang Zhou}, {and} \bibinfo{person}{Yang Wang}.} \bibinfo{year}{2024}\natexlab{}.
\newblock \showarticletitle{HDMixer: Hierarchical Dependency with Extendable Patch for Multivariate Time Series Forecasting}. In \bibinfo{booktitle}{\emph{AAAI}}. \bibinfo{pages}{12608--12616}.
\newblock


\bibitem[Jhamtani and Berg-Kirkpatrick(2021)]%
        {jhamtani-berg-kirkpatrick-2021-truth}
\bibfield{author}{\bibinfo{person}{Harsh Jhamtani} {and} \bibinfo{person}{Taylor Berg-Kirkpatrick}.} \bibinfo{year}{2021}\natexlab{}.
\newblock \showarticletitle{Truth-Conditional Captions for Time Series Data}. In \bibinfo{booktitle}{\emph{Proceedings of the 2021 Conference on Empirical Methods in Natural Language Processing}}, \bibfield{editor}{\bibinfo{person}{Marie-Francine Moens}, \bibinfo{person}{Xuanjing Huang}, \bibinfo{person}{Lucia Specia}, {and} \bibinfo{person}{Scott Wen-tau Yih}} (Eds.). \bibinfo{publisher}{Association for Computational Linguistics}, \bibinfo{address}{Online and Punta Cana, Dominican Republic}, \bibinfo{pages}{719--733}.
\newblock
\urldef\tempurl%
\url{https://doi.org/10.18653/v1/2021.emnlp-main.55}
\showDOI{\tempurl}


\bibitem[Jia et~al\mbox{.}(2024)]%
        {gpt4ms}
\bibfield{author}{\bibinfo{person}{Furong Jia}, \bibinfo{person}{Kevin Wang}, \bibinfo{person}{Yixiang Zheng}, \bibinfo{person}{Defu Cao}, {and} \bibinfo{person}{Yan Liu}.} \bibinfo{year}{2024}\natexlab{}.
\newblock \showarticletitle{GPT4MTS: Prompt-based Large Language Model for Multimodal Time-series Forecasting}.
\newblock \bibinfo{journal}{\emph{Proceedings of the AAAI Conference on Artificial Intelligence}} \bibinfo{volume}{38}, \bibinfo{number}{21} (\bibinfo{date}{Mar.} \bibinfo{year}{2024}), \bibinfo{pages}{23343--23351}.
\newblock
\urldef\tempurl%
\url{https://doi.org/10.1609/aaai.v38i21.30383}
\showDOI{\tempurl}


\bibitem[Jin et~al\mbox{.}(2024)]%
        {timellm}
\bibfield{author}{\bibinfo{person}{Ming Jin}, \bibinfo{person}{Shiyu Wang}, \bibinfo{person}{Lintao Ma}, \bibinfo{person}{Zhixuan Chu}, \bibinfo{person}{James~Y. Zhang}, \bibinfo{person}{Xiaoming Shi}, \bibinfo{person}{Pin-Yu Chen}, \bibinfo{person}{Yuxuan Liang}, \bibinfo{person}{Yuan-Fang Li}, \bibinfo{person}{Shirui Pan}, {and} \bibinfo{person}{Qingsong Wen}.} \bibinfo{year}{2024}\natexlab{}.
\newblock \showarticletitle{Time-{LLM}: Time Series Forecasting by Reprogramming Large Language Models}. In \bibinfo{booktitle}{\emph{The Twelfth International Conference on Learning Representations}}.
\newblock


\bibitem[Kudrat et~al\mbox{.}(2025)]%
        {kudrat2025patchwisestructurallosstime}
\bibfield{author}{\bibinfo{person}{Dilfira Kudrat}, \bibinfo{person}{Zongxia Xie}, \bibinfo{person}{Yanru Sun}, \bibinfo{person}{Tianyu Jia}, {and} \bibinfo{person}{Qinghua Hu}.} \bibinfo{year}{2025}\natexlab{}.
\newblock \bibinfo{title}{Patch-wise Structural Loss for Time Series Forecasting}.
\newblock
\newblock
\showeprint[arxiv]{2503.00877}~[cs.LG]


\bibitem[Lee et~al\mbox{.}(2023)]%
        {lee2023vectorquantizedtimeseries}
\bibfield{author}{\bibinfo{person}{Daesoo Lee}, \bibinfo{person}{Sara Malacarne}, {and} \bibinfo{person}{Erlend Aune}.} \bibinfo{year}{2023}\natexlab{}.
\newblock \bibinfo{title}{Vector Quantized Time Series Generation with a Bidirectional Prior Model}.
\newblock
\newblock
\showeprint[arxiv]{2303.04743}~[cs.LG]


\bibitem[Lin et~al\mbox{.}(2024)]%
        {lin2024decodingtimeseriesllms}
\bibfield{author}{\bibinfo{person}{Minhua Lin}, \bibinfo{person}{Zhengzhang Chen}, \bibinfo{person}{Yanchi Liu}, \bibinfo{person}{Xujiang Zhao}, \bibinfo{person}{Zongyu Wu}, \bibinfo{person}{Junxiang Wang}, \bibinfo{person}{Xiang Zhang}, \bibinfo{person}{Suhang Wang}, {and} \bibinfo{person}{Haifeng Chen}.} \bibinfo{year}{2024}\natexlab{}.
\newblock \bibinfo{title}{Decoding Time Series with LLMs: A Multi-Agent Framework for Cross-Domain Annotation}.
\newblock
\newblock
\showeprint[arxiv]{2410.17462}~[cs.AI]


\bibitem[Liu et~al\mbox{.}(2025)]%
        {timecma}
\bibfield{author}{\bibinfo{person}{Chenxi Liu}, \bibinfo{person}{Qianxiong Xu}, \bibinfo{person}{Hao Miao}, \bibinfo{person}{Sun Yang}, \bibinfo{person}{Lingzheng Zhang}, \bibinfo{person}{Cheng Long}, \bibinfo{person}{Ziyue Li}, {and} \bibinfo{person}{Rui Zhao}.} \bibinfo{year}{2025}\natexlab{}.
\newblock \showarticletitle{{TimeCMA}: Towards LLM-Empowered Multivariate Time Series Forecasting via Cross-Modality Alignment}. In \bibinfo{booktitle}{\emph{AAAI}}.
\newblock


\bibitem[Liu et~al\mbox{.}(2024g)]%
        {timemmd}
\bibfield{author}{\bibinfo{person}{Haoxin Liu}, \bibinfo{person}{Shangqing Xu}, \bibinfo{person}{Zhiyuan Zhao}, \bibinfo{person}{Lingkai Kong}, \bibinfo{person}{Harshavardhan Kamarthi}, \bibinfo{person}{Aditya~B. Sasanur}, \bibinfo{person}{Megha Sharma}, \bibinfo{person}{Jiaming Cui}, \bibinfo{person}{Qingsong Wen}, \bibinfo{person}{Chao Zhang}, {and} \bibinfo{person}{B.~Aditya Prakash}.} \bibinfo{year}{2024}\natexlab{g}.
\newblock \showarticletitle{Time-{MMD}: Multi-Domain Multimodal Dataset for Time Series Analysis}. In \bibinfo{booktitle}{\emph{The Thirty-eight Conference on Neural Information Processing Systems Datasets and Benchmarks Track}}.
\newblock


\bibitem[Liu et~al\mbox{.}(2024b)]%
        {calf}
\bibfield{author}{\bibinfo{person}{Peiyuan Liu}, \bibinfo{person}{Hang Guo}, \bibinfo{person}{Tao Dai}, \bibinfo{person}{Naiqi Li}, \bibinfo{person}{Jigang Bao}, \bibinfo{person}{Xudong Ren}, \bibinfo{person}{Yong Jiang}, {and} \bibinfo{person}{Shu-Tao Xia}.} \bibinfo{year}{2024}\natexlab{b}.
\newblock \showarticletitle{CALF: Aligning LLMs for Time Series Forecasting via Cross-modal Fine-Tuning}.
\newblock \bibinfo{journal}{\emph{arXiv preprint arXiv:2403.07300}} (\bibinfo{year}{2024}).
\newblock


\bibitem[Liu et~al\mbox{.}(2024a)]%
        {liu2024dgcformerdeepgraphclustering}
\bibfield{author}{\bibinfo{person}{Qinshuo Liu}, \bibinfo{person}{Yanwen Fang}, \bibinfo{person}{Pengtao Jiang}, {and} \bibinfo{person}{Guodong Li}.} \bibinfo{year}{2024}\natexlab{a}.
\newblock \bibinfo{title}{DGCformer: Deep Graph Clustering Transformer for Multivariate Time Series Forecasting}.
\newblock
\newblock
\showeprint[arxiv]{2405.08440}~[cs.LG]


\bibitem[Liu et~al\mbox{.}(2024e)]%
        {liu2024timeffm}
\bibfield{author}{\bibinfo{person}{Qingxiang Liu}, \bibinfo{person}{Xu Liu}, \bibinfo{person}{Chenghao Liu}, \bibinfo{person}{Qingsong Wen}, {and} \bibinfo{person}{Yuxuan Liang}.} \bibinfo{year}{2024}\natexlab{e}.
\newblock \showarticletitle{Time-{FFM}: Towards {LM}-Empowered Federated Foundation Model for Time Series Forecasting}. In \bibinfo{booktitle}{\emph{The Thirty-eighth Annual Conference on Neural Information Processing Systems}}.
\newblock


\bibitem[Liu et~al\mbox{.}(2024c)]%
        {liu2024unitime}
\bibfield{author}{\bibinfo{person}{Xu Liu}, \bibinfo{person}{Junfeng Hu}, \bibinfo{person}{Yuan Li}, \bibinfo{person}{Shizhe Diao}, \bibinfo{person}{Yuxuan Liang}, \bibinfo{person}{Bryan Hooi}, {and} \bibinfo{person}{Roger Zimmermann}.} \bibinfo{year}{2024}\natexlab{c}.
\newblock \showarticletitle{UniTime: A Language-Empowered Unified Model for Cross-Domain Time Series Forecasting}. In \bibinfo{booktitle}{\emph{Proceedings of the ACM Web Conference 2024}}.
\newblock


\bibitem[Liu et~al\mbox{.}(2024d)]%
        {liu2024itransformer}
\bibfield{author}{\bibinfo{person}{Yong Liu}, \bibinfo{person}{Tengge Hu}, \bibinfo{person}{Haoran Zhang}, \bibinfo{person}{Haixu Wu}, \bibinfo{person}{Shiyu Wang}, \bibinfo{person}{Lintao Ma}, {and} \bibinfo{person}{Mingsheng Long}.} \bibinfo{year}{2024}\natexlab{d}.
\newblock \showarticletitle{iTransformer: Inverted Transformers Are Effective for Time Series Forecasting}. In \bibinfo{booktitle}{\emph{The Twelfth International Conference on Learning Representations}}.
\newblock


\bibitem[Liu et~al\mbox{.}(2024f)]%
        {liu2024autotimes}
\bibfield{author}{\bibinfo{person}{Yong Liu}, \bibinfo{person}{Guo Qin}, \bibinfo{person}{Xiangdong Huang}, \bibinfo{person}{Jianmin Wang}, {and} \bibinfo{person}{Mingsheng Long}.} \bibinfo{year}{2024}\natexlab{f}.
\newblock \showarticletitle{AutoTimes: Autoregressive Time Series Forecasters via Large Language Models}. In \bibinfo{booktitle}{\emph{The Thirty-eighth Annual Conference on Neural Information Processing Systems}}.
\newblock


\bibitem[Liu et~al\mbox{.}(2022)]%
        {liu2022non}
\bibfield{author}{\bibinfo{person}{Yong Liu}, \bibinfo{person}{Haixu Wu}, \bibinfo{person}{Jianmin Wang}, {and} \bibinfo{person}{Mingsheng Long}.} \bibinfo{year}{2022}\natexlab{}.
\newblock \showarticletitle{Non-stationary Transformers: Exploring the Stationarity in Time Series Forecasting}.
\newblock  (\bibinfo{year}{2022}).
\newblock


\bibitem[Mahinpei et~al\mbox{.}(2022)]%
        {lincap}
\bibfield{author}{\bibinfo{person}{Anita Mahinpei}, \bibinfo{person}{Zona Kostic}, {and} \bibinfo{person}{Chris Tanner}.} \bibinfo{year}{2022}\natexlab{}.
\newblock \showarticletitle{LineCap: Line Charts for Data Visualization Captioning Models}. In \bibinfo{booktitle}{\emph{2022 IEEE Visualization and Visual Analytics (VIS)}}. \bibinfo{pages}{35--39}.
\newblock
\urldef\tempurl%
\url{https://doi.org/10.1109/VIS54862.2022.00016}
\showDOI{\tempurl}


\bibitem[Meng et~al\mbox{.}(2024)]%
        {meng-etal-2024-chartassistant}
\bibfield{author}{\bibinfo{person}{Fanqing Meng}, \bibinfo{person}{Wenqi Shao}, \bibinfo{person}{Quanfeng Lu}, \bibinfo{person}{Peng Gao}, \bibinfo{person}{Kaipeng Zhang}, \bibinfo{person}{Yu Qiao}, {and} \bibinfo{person}{Ping Luo}.} \bibinfo{year}{2024}\natexlab{}.
\newblock \showarticletitle{{C}hart{A}ssistant: A Universal Chart Multimodal Language Model via Chart-to-Table Pre-training and Multitask Instruction Tuning}. In \bibinfo{booktitle}{\emph{Findings of the Association for Computational Linguistics: ACL 2024}}, \bibfield{editor}{\bibinfo{person}{Lun-Wei Ku}, \bibinfo{person}{Andre Martins}, {and} \bibinfo{person}{Vivek Srikumar}} (Eds.). \bibinfo{publisher}{Association for Computational Linguistics}, \bibinfo{address}{Bangkok, Thailand}, \bibinfo{pages}{7775--7803}.
\newblock
\urldef\tempurl%
\url{https://doi.org/10.18653/v1/2024.findings-acl.463}
\showDOI{\tempurl}


\bibitem[Merrill et~al\mbox{.}(2024)]%
        {merrill}
\bibfield{author}{\bibinfo{person}{Mike~A Merrill}, \bibinfo{person}{Mingtian Tan}, \bibinfo{person}{Vinayak Gupta}, \bibinfo{person}{Thomas Hartvigsen}, {and} \bibinfo{person}{Tim Althoff}.} \bibinfo{year}{2024}\natexlab{}.
\newblock \showarticletitle{Language Models Still Struggle to Zero-shot Reason about Time Series}. In \bibinfo{booktitle}{\emph{Findings of the Association for Computational Linguistics: EMNLP 2024}}, \bibfield{editor}{\bibinfo{person}{Yaser Al-Onaizan}, \bibinfo{person}{Mohit Bansal}, {and} \bibinfo{person}{Yun-Nung Chen}} (Eds.). \bibinfo{publisher}{Association for Computational Linguistics}, \bibinfo{address}{Miami, Florida, USA}, \bibinfo{pages}{3512--3533}.
\newblock
\urldef\tempurl%
\url{https://doi.org/10.18653/v1/2024.findings-emnlp.201}
\showDOI{\tempurl}


\bibitem[Miao et~al\mbox{.}(2024)]%
        {miao2024}
\bibfield{author}{\bibinfo{person}{Hao Miao}, \bibinfo{person}{Yan Zhao}, \bibinfo{person}{Chenjuan Guo}, \bibinfo{person}{Bin Yang}, \bibinfo{person}{Kai Zheng}, \bibinfo{person}{Feiteng Huang}, \bibinfo{person}{Jiandong Xie}, {and} \bibinfo{person}{Christian~S. Jensen}.} \bibinfo{year}{2024}\natexlab{}.
\newblock \showarticletitle{A Unified Replay-Based Continuous Learning Framework for Spatio-Temporal Prediction on Streaming Data}. In \bibinfo{booktitle}{\emph{2024 IEEE 40th International Conference on Data Engineering (ICDE)}}. \bibinfo{pages}{1050--1062}.
\newblock
\urldef\tempurl%
\url{https://doi.org/10.1109/ICDE60146.2024.00085}
\showDOI{\tempurl}


\bibitem[Murakami et~al\mbox{.}(2017)]%
        {murakami-etal-2017-learning}
\bibfield{author}{\bibinfo{person}{Soichiro Murakami}, \bibinfo{person}{Akihiko Watanabe}, \bibinfo{person}{Akira Miyazawa}, \bibinfo{person}{Keiichi Goshima}, \bibinfo{person}{Toshihiko Yanase}, \bibinfo{person}{Hiroya Takamura}, {and} \bibinfo{person}{Yusuke Miyao}.} \bibinfo{year}{2017}\natexlab{}.
\newblock \showarticletitle{Learning to Generate Market Comments from Stock Prices}. In \bibinfo{booktitle}{\emph{Proceedings of the 55th Annual Meeting of the Association for Computational Linguistics (Volume 1: Long Papers)}}, \bibfield{editor}{\bibinfo{person}{Regina Barzilay} {and} \bibinfo{person}{Min-Yen Kan}} (Eds.). \bibinfo{publisher}{Association for Computational Linguistics}, \bibinfo{address}{Vancouver, Canada}, \bibinfo{pages}{1374--1384}.
\newblock
\urldef\tempurl%
\url{https://doi.org/10.18653/v1/P17-1126}
\showDOI{\tempurl}


\bibitem[Nie et~al\mbox{.}(2023)]%
        {patchtst}
\bibfield{author}{\bibinfo{person}{Yuqi Nie}, \bibinfo{person}{Nam~H Nguyen}, \bibinfo{person}{Phanwadee Sinthong}, {and} \bibinfo{person}{Jayant Kalagnanam}.} \bibinfo{year}{2023}\natexlab{}.
\newblock \showarticletitle{A Time Series is Worth 64 Words: Long-term Forecasting with Transformers}. In \bibinfo{booktitle}{\emph{The Eleventh International Conference on Learning Representations}}.
\newblock


\bibitem[Pan et~al\mbox{.}(2024)]%
        {s2ip}
\bibfield{author}{\bibinfo{person}{Zijie Pan}, \bibinfo{person}{Yushan Jiang}, \bibinfo{person}{Sahil Garg}, \bibinfo{person}{Anderson Schneider}, \bibinfo{person}{Yuriy Nevmyvaka}, {and} \bibinfo{person}{Dongjin Song}.} \bibinfo{year}{2024}\natexlab{}.
\newblock \showarticletitle{\$S{\textasciicircum}2\${IP}-{LLM}: Semantic Space Informed Prompt Learning with {LLM} for Time Series Forecasting}. In \bibinfo{booktitle}{\emph{Forty-first International Conference on Machine Learning}}.
\newblock


\bibitem[Peng et~al\mbox{.}(2023)]%
        {peng2023}
\bibfield{author}{\bibinfo{person}{Cheng Peng}, \bibinfo{person}{Xi Yang}, \bibinfo{person}{Aokun Chen}, \bibinfo{person}{Kaleb~E. Smith}, \bibinfo{person}{Nima PourNejatian}, \bibinfo{person}{Anthony~B. Costa}, \bibinfo{person}{Cheryl Martin}, \bibinfo{person}{Mona~G. Flores}, \bibinfo{person}{Ying Zhang}, \bibinfo{person}{Tanja Magoc}, \bibinfo{person}{Gloria Lipori}, \bibinfo{person}{Duane~A. Mitchell}, \bibinfo{person}{Naykky~S. Ospina}, \bibinfo{person}{Mustafa~M. Ahmed}, \bibinfo{person}{William~R. Hogan}, \bibinfo{person}{Elizabeth~A. Shenkman}, \bibinfo{person}{Yi Guo}, \bibinfo{person}{Jiang Bian}, {and} \bibinfo{person}{Yonghui Wu}.} \bibinfo{year}{2023}\natexlab{}.
\newblock \showarticletitle{A study of generative large language model for medical research and healthcare}.
\newblock \bibinfo{journal}{\emph{npj Digital Medicine}} \bibinfo{volume}{6}, \bibinfo{number}{210} (\bibinfo{year}{2023}), \bibinfo{pages}{210}.
\newblock
\urldef\tempurl%
\url{https://doi.org/10.1038/s41746-023-00958-w}
\showDOI{\tempurl}


\bibitem[Qiu et~al\mbox{.}(2024)]%
        {qiuTSF}
\bibfield{author}{\bibinfo{person}{Xiangfei Qiu}, \bibinfo{person}{Jilin Hu}, \bibinfo{person}{Lekui Zhou}, \bibinfo{person}{Xingjian Wu}, \bibinfo{person}{Junyang Du}, \bibinfo{person}{Buang Zhang}, \bibinfo{person}{Chenjuan Guo}, \bibinfo{person}{Aoying Zhou}, \bibinfo{person}{Christian~S. Jensen}, \bibinfo{person}{Zhenli Sheng}, {and} \bibinfo{person}{Bin Yang}.} \bibinfo{year}{2024}\natexlab{}.
\newblock \showarticletitle{TFB: Towards Comprehensive and Fair Benchmarking of Time Series Forecasting Methods}.
\newblock \bibinfo{journal}{\emph{Proc. VLDB Endow.}} \bibinfo{volume}{17}, \bibinfo{number}{9} (\bibinfo{date}{May} \bibinfo{year}{2024}), \bibinfo{pages}{2363–2377}.
\newblock
\showISSN{2150-8097}
\urldef\tempurl%
\url{https://doi.org/10.14778/3665844.3665863}
\showDOI{\tempurl}


\bibitem[Qiu et~al\mbox{.}(2025)]%
        {qiu2025duet}
\bibfield{author}{\bibinfo{person}{Xiangfei Qiu}, \bibinfo{person}{Xingjian Wu}, \bibinfo{person}{Yan Lin}, \bibinfo{person}{Chenjuan Guo}, \bibinfo{person}{Jilin Hu}, {and} \bibinfo{person}{Bin Yang}.} \bibinfo{year}{2025}\natexlab{}.
\newblock \showarticletitle{DUET: Dual Clustering Enhanced Multivariate Time Series Forecasting}. In \bibinfo{booktitle}{\emph{SIGKDD}}.
\newblock


\bibitem[Shao et~al\mbox{.}(2024)]%
        {shao2024exploring}
\bibfield{author}{\bibinfo{person}{Zezhi Shao}, \bibinfo{person}{Fei Wang}, \bibinfo{person}{Yongjun Xu}, \bibinfo{person}{Wei Wei}, \bibinfo{person}{Chengqing Yu}, \bibinfo{person}{Zhao Zhang}, \bibinfo{person}{Di Yao}, \bibinfo{person}{Tao Sun}, \bibinfo{person}{Guangyin Jin}, \bibinfo{person}{Xin Cao}, {et~al\mbox{.}}} \bibinfo{year}{2024}\natexlab{}.
\newblock \showarticletitle{Exploring progress in multivariate time series forecasting: Comprehensive benchmarking and heterogeneity analysis}.
\newblock \bibinfo{journal}{\emph{IEEE Transactions on Knowledge and Data Engineering}} \bibinfo{volume}{37}, \bibinfo{number}{1} (\bibinfo{year}{2024}), \bibinfo{pages}{291--305}.
\newblock


\bibitem[Shi et~al\mbox{.}(2025)]%
        {shi2025timemoe}
\bibfield{author}{\bibinfo{person}{Xiaoming Shi}, \bibinfo{person}{Shiyu Wang}, \bibinfo{person}{Yuqi Nie}, \bibinfo{person}{Dianqi Li}, \bibinfo{person}{Zhou Ye}, \bibinfo{person}{Qingsong Wen}, {and} \bibinfo{person}{Ming Jin}.} \bibinfo{year}{2025}\natexlab{}.
\newblock \showarticletitle{Time-MoE: Billion-Scale Time Series Foundation Models with Mixture of Experts}. In \bibinfo{booktitle}{\emph{The Thirteenth International Conference on Learning Representations}}.
\newblock


\bibitem[Smith and Demetsky(1997)]%
        {brain1997}
\bibfield{author}{\bibinfo{person}{Brian~L. Smith} {and} \bibinfo{person}{Michael~J. Demetsky}.} \bibinfo{year}{1997}\natexlab{}.
\newblock \showarticletitle{Traffic Flow Forecasting: Comparison of Modeling Approaches}.
\newblock \bibinfo{journal}{\emph{Journal of Transportation Engineering}} \bibinfo{volume}{123}, \bibinfo{number}{4} (\bibinfo{year}{1997}), \bibinfo{pages}{261--266}.
\newblock
\urldef\tempurl%
\url{https://doi.org/10.1061/(ASCE)0733-947X(1997)123:4(261)}
\showDOI{\tempurl}


\bibitem[Sowdaboina et~al\mbox{.}(2014)]%
        {sumts}
\bibfield{author}{\bibinfo{person}{Pranay Kumar~Venkata Sowdaboina}, \bibinfo{person}{Sutanu Chakraborti}, {and} \bibinfo{person}{Somayajulu Sripada}.} \bibinfo{year}{2014}\natexlab{}.
\newblock \showarticletitle{Learning to Summarize Time Series Data}. In \bibinfo{booktitle}{\emph{Computational Linguistics and Intelligent Text Processing}}, \bibfield{editor}{\bibinfo{person}{Alexander Gelbukh}} (Ed.). \bibinfo{publisher}{Springer Berlin Heidelberg}, \bibinfo{address}{Berlin, Heidelberg}, \bibinfo{pages}{515--528}.
\newblock
\showISBNx{978-3-642-54906-9}


\bibitem[Su et~al\mbox{.}(2019)]%
        {10.1145/3292500.3330672}
\bibfield{author}{\bibinfo{person}{Ya Su}, \bibinfo{person}{Youjian Zhao}, \bibinfo{person}{Chenhao Niu}, \bibinfo{person}{Rong Liu}, \bibinfo{person}{Wei Sun}, {and} \bibinfo{person}{Dan Pei}.} \bibinfo{year}{2019}\natexlab{}.
\newblock \showarticletitle{Robust Anomaly Detection for Multivariate Time Series through Stochastic Recurrent Neural Network}. In \bibinfo{booktitle}{\emph{Proceedings of the 25th ACM SIGKDD International Conference on Knowledge Discovery \& Data Mining}} (Anchorage, AK, USA) \emph{(\bibinfo{series}{KDD '19})}. \bibinfo{publisher}{Association for Computing Machinery}, \bibinfo{address}{New York, NY, USA}, \bibinfo{pages}{2828–2837}.
\newblock
\showISBNx{9781450362016}
\urldef\tempurl%
\url{https://doi.org/10.1145/3292500.3330672}
\showDOI{\tempurl}


\bibitem[Sun et~al\mbox{.}(2024)]%
        {test}
\bibfield{author}{\bibinfo{person}{Chenxi Sun}, \bibinfo{person}{Hongyan Li}, \bibinfo{person}{Yaliang Li}, {and} \bibinfo{person}{Shenda Hong}.} \bibinfo{year}{2024}\natexlab{}.
\newblock \showarticletitle{{TEST}: Text Prototype Aligned Embedding to Activate {LLM}'s Ability for Time Series}. In \bibinfo{booktitle}{\emph{The Twelfth International Conference on Learning Representations}}.
\newblock


\bibitem[Tan et~al\mbox{.}(2024)]%
        {tan2024are}
\bibfield{author}{\bibinfo{person}{Mingtian Tan}, \bibinfo{person}{Mike~A Merrill}, \bibinfo{person}{Vinayak Gupta}, \bibinfo{person}{Tim Althoff}, {and} \bibinfo{person}{Thomas Hartvigsen}.} \bibinfo{year}{2024}\natexlab{}.
\newblock \showarticletitle{Are Language Models Actually Useful for Time Series Forecasting?}. In \bibinfo{booktitle}{\emph{The Thirty-eighth Annual Conference on Neural Information Processing Systems}}.
\newblock


\bibitem[Tang et~al\mbox{.}(2025b)]%
        {tsfl}
\bibfield{author}{\bibinfo{person}{Hua Tang}, \bibinfo{person}{Chong Zhang}, \bibinfo{person}{Mingyu Jin}, \bibinfo{person}{Qinkai Yu}, \bibinfo{person}{Zhenting Wang}, \bibinfo{person}{Xiaobo Jin}, \bibinfo{person}{Yongfeng Zhang}, {and} \bibinfo{person}{Mengnan Du}.} \bibinfo{year}{2025}\natexlab{b}.
\newblock \showarticletitle{Time Series Forecasting with LLMs: Understanding and Enhancing Model Capabilities}.
\newblock \bibinfo{journal}{\emph{SIGKDD Explor. Newsl.}} \bibinfo{volume}{26}, \bibinfo{number}{2} (\bibinfo{date}{Jan.} \bibinfo{year}{2025}), \bibinfo{pages}{109–118}.
\newblock
\showISSN{1931-0145}
\urldef\tempurl%
\url{https://doi.org/10.1145/3715073.3715083}
\showDOI{\tempurl}


\bibitem[Tang et~al\mbox{.}(2025a)]%
        {llmps}
\bibfield{author}{\bibinfo{person}{Jialiang Tang}, \bibinfo{person}{Shuo Chen}, \bibinfo{person}{Chen Gong}, \bibinfo{person}{Jing Zhang}, {and} \bibinfo{person}{Dacheng Tao}.} \bibinfo{year}{2025}\natexlab{a}.
\newblock \bibinfo{title}{LLM-PS: Empowering Large Language Models for Time Series Forecasting with Temporal Patterns and Semantics}.
\newblock
\newblock
\showeprint[arxiv]{2503.09656}~[cs.LG]


\bibitem[Trabelsi et~al\mbox{.}(2025)]%
        {trabelsi2025timeserieslanguagemodel}
\bibfield{author}{\bibinfo{person}{Mohamed Trabelsi}, \bibinfo{person}{Aidan Boyd}, \bibinfo{person}{Jin Cao}, {and} \bibinfo{person}{Huseyin Uzunalioglu}.} \bibinfo{year}{2025}\natexlab{}.
\newblock \bibinfo{title}{Time Series Language Model for Descriptive Caption Generation}.
\newblock
\newblock
\showeprint[arxiv]{2501.01832}~[cs.CL]


\bibitem[van Panhuis et~al\mbox{.}(2018)]%
        {van2023}
\bibfield{author}{\bibinfo{person}{Willem~G van Panhuis}, \bibinfo{person}{Anne Cross}, {and} \bibinfo{person}{Donald~S Burke}.} \bibinfo{year}{2018}\natexlab{}.
\newblock \showarticletitle{Project Tycho 2.0: a repository to improve the integration and reuse of data for global population health}.
\newblock \bibinfo{journal}{\emph{Journal of the American Medical Informatics Association}} \bibinfo{volume}{25}, \bibinfo{number}{12} (\bibinfo{date}{10} \bibinfo{year}{2018}), \bibinfo{pages}{1608--1617}.
\newblock
\showISSN{1527-974X}
\urldef\tempurl%
\url{https://doi.org/10.1093/jamia/ocy123}
\showDOI{\tempurl}


\bibitem[Wang et~al\mbox{.}(2025)]%
        {ChatTime}
\bibfield{author}{\bibinfo{person}{Chengsen Wang}, \bibinfo{person}{Qi Qi}, \bibinfo{person}{Jingyu Wang}, \bibinfo{person}{Haifeng Sun}, \bibinfo{person}{Zirui Zhuang}, \bibinfo{person}{Jinming Wu}, \bibinfo{person}{Lei Zhang}, {and} \bibinfo{person}{Jianxin Liao}.} \bibinfo{year}{2025}\natexlab{}.
\newblock \showarticletitle{ChatTime: A Unified Multimodal Time Series Foundation Model Bridging Numerical and Textual Data}. In \bibinfo{booktitle}{\emph{AAAI Conference on Artificial Intelligence}}.
\newblock


\bibitem[Wang et~al\mbox{.}(2023b)]%
        {wang2023micn}
\bibfield{author}{\bibinfo{person}{Huiqiang Wang}, \bibinfo{person}{Jian Peng}, \bibinfo{person}{Feihu Huang}, \bibinfo{person}{Jince Wang}, \bibinfo{person}{Junhui Chen}, {and} \bibinfo{person}{Yifei Xiao}.} \bibinfo{year}{2023}\natexlab{b}.
\newblock \showarticletitle{{MICN}: Multi-scale Local and Global Context Modeling for Long-term Series Forecasting}. In \bibinfo{booktitle}{\emph{The Eleventh International Conference on Learning Representations}}.
\newblock


\bibitem[Wang et~al\mbox{.}(2024b)]%
        {wang2024timemixer}
\bibfield{author}{\bibinfo{person}{Shiyu Wang}, \bibinfo{person}{Haixu Wu}, \bibinfo{person}{Xiaoming Shi}, \bibinfo{person}{Tengge Hu}, \bibinfo{person}{Huakun Luo}, \bibinfo{person}{Lintao Ma}, \bibinfo{person}{James~Y. Zhang}, {and} \bibinfo{person}{JUN ZHOU}.} \bibinfo{year}{2024}\natexlab{b}.
\newblock \showarticletitle{TimeMixer: Decomposable Multiscale Mixing for Time Series Forecasting}. In \bibinfo{booktitle}{\emph{The Twelfth International Conference on Learning Representations}}.
\newblock


\bibitem[Wang et~al\mbox{.}(2023a)]%
        {WANG2023110214}
\bibfield{author}{\bibinfo{person}{Xingyu Wang}, \bibinfo{person}{Hui Liu}, \bibinfo{person}{Junzhao Du}, \bibinfo{person}{Xiyao Dong}, {and} \bibinfo{person}{Zhihan Yang}.} \bibinfo{year}{2023}\natexlab{a}.
\newblock \showarticletitle{A long-term multivariate time series forecasting network combining series decomposition and convolutional neural networks}.
\newblock \bibinfo{journal}{\emph{Applied Soft Computing}}  \bibinfo{volume}{139} (\bibinfo{year}{2023}), \bibinfo{pages}{110214}.
\newblock
\showISSN{1568-4946}
\urldef\tempurl%
\url{https://doi.org/10.1016/j.asoc.2023.110214}
\showDOI{\tempurl}


\bibitem[Wang et~al\mbox{.}(2024a)]%
        {wang2024medformer}
\bibfield{author}{\bibinfo{person}{Yihe Wang}, \bibinfo{person}{Nan Huang}, \bibinfo{person}{Taida Li}, \bibinfo{person}{Yujun Yan}, {and} \bibinfo{person}{Xiang Zhang}.} \bibinfo{year}{2024}\natexlab{a}.
\newblock \showarticletitle{Medformer: A Multi-Granularity Patching Transformer for Medical Time-Series Classification}. In \bibinfo{booktitle}{\emph{The Thirty-eighth Annual Conference on Neural Information Processing Systems}}.
\newblock


\bibitem[Wu et~al\mbox{.}(2023a)]%
        {timesnet}
\bibfield{author}{\bibinfo{person}{Haixu Wu}, \bibinfo{person}{Tengge Hu}, \bibinfo{person}{Yong Liu}, \bibinfo{person}{Hang Zhou}, \bibinfo{person}{Jianmin Wang}, {and} \bibinfo{person}{Mingsheng Long}.} \bibinfo{year}{2023}\natexlab{a}.
\newblock \showarticletitle{TimesNet: Temporal 2D-Variation Modeling for General Time Series Analysis}. In \bibinfo{booktitle}{\emph{International Conference on Learning Representations}}.
\newblock


\bibitem[Wu et~al\mbox{.}(2021)]%
        {wu2021autoformer}
\bibfield{author}{\bibinfo{person}{Haixu Wu}, \bibinfo{person}{Jiehui Xu}, \bibinfo{person}{Jianmin Wang}, {and} \bibinfo{person}{Mingsheng Long}.} \bibinfo{year}{2021}\natexlab{}.
\newblock \showarticletitle{Autoformer: Decomposition Transformers with Auto-Correlation for Long-Term Series Forecasting}. In \bibinfo{booktitle}{\emph{Advances in Neural Information Processing Systems}}, \bibfield{editor}{\bibinfo{person}{A.~Beygelzimer}, \bibinfo{person}{Y.~Dauphin}, \bibinfo{person}{P.~Liang}, {and} \bibinfo{person}{J.~Wortman Vaughan}} (Eds.).
\newblock


\bibitem[Wu et~al\mbox{.}(2023b)]%
        {wu2023}
\bibfield{author}{\bibinfo{person}{Haixu Wu}, \bibinfo{person}{Hang Zhou}, \bibinfo{person}{Mingsheng Long}, {and} \bibinfo{person}{Jianmin Wang}.} \bibinfo{year}{2023}\natexlab{b}.
\newblock \showarticletitle{Interpretable weather forecasting for worldwide stations with a unified deep model}.
\newblock \bibinfo{journal}{\emph{Nat. Mac. Intell.}} \bibinfo{volume}{5}, \bibinfo{number}{6} (\bibinfo{date}{June} \bibinfo{year}{2023}), \bibinfo{pages}{602--611}.
\newblock


\bibitem[Yu et~al\mbox{.}(2024)]%
        {yu2024ginar}
\bibfield{author}{\bibinfo{person}{Chengqing Yu}, \bibinfo{person}{Fei Wang}, \bibinfo{person}{Zezhi Shao}, \bibinfo{person}{Tangwen Qian}, \bibinfo{person}{Zhao Zhang}, \bibinfo{person}{Wei Wei}, {and} \bibinfo{person}{Yongjun Xu}.} \bibinfo{year}{2024}\natexlab{}.
\newblock \showarticletitle{Ginar: An end-to-end multivariate time series forecasting model suitable for variable missing}. In \bibinfo{booktitle}{\emph{SIGKDD}}. \bibinfo{pages}{3989--4000}.
\newblock


\bibitem[Yu et~al\mbox{.}(2023)]%
        {DSformer}
\bibfield{author}{\bibinfo{person}{Chengqing Yu}, \bibinfo{person}{Fei Wang}, \bibinfo{person}{Zezhi Shao}, \bibinfo{person}{Tao Sun}, \bibinfo{person}{Lin Wu}, {and} \bibinfo{person}{Yongjun Xu}.} \bibinfo{year}{2023}\natexlab{}.
\newblock \showarticletitle{DSformer: {A} Double Sampling Transformer for Multivariate Time Series Long-term Prediction}. In \bibinfo{booktitle}{\emph{CIKM}}. \bibinfo{pages}{3062--3072}.
\newblock


\bibitem[Zeng et~al\mbox{.}(2023)]%
        {dlinear}
\bibfield{author}{\bibinfo{person}{Ailing Zeng}, \bibinfo{person}{Muxi Chen}, \bibinfo{person}{Lei Zhang}, {and} \bibinfo{person}{Qiang Xu}.} \bibinfo{year}{2023}\natexlab{}.
\newblock \showarticletitle{Are transformers effective for time series forecasting?}. In \bibinfo{booktitle}{\emph{Proceedings of the Thirty-Seventh AAAI Conference on Artificial Intelligence and Thirty-Fifth Conference on Innovative Applications of Artificial Intelligence and Thirteenth Symposium on Educational Advances in Artificial Intelligence}} \emph{(\bibinfo{series}{AAAI'23/IAAI'23/EAAI'23})}. \bibinfo{publisher}{AAAI Press}, Article \bibinfo{articleno}{1248}, \bibinfo{numpages}{8}~pages.
\newblock
\showISBNx{978-1-57735-880-0}
\urldef\tempurl%
\url{https://doi.org/10.1609/aaai.v37i9.26317}
\showDOI{\tempurl}


\bibitem[Zhang et~al\mbox{.}(2024)]%
        {zhang2024not}
\bibfield{author}{\bibinfo{person}{Xingyu Zhang}, \bibinfo{person}{Siyu Zhao}, \bibinfo{person}{Zeen Song}, \bibinfo{person}{Huijie Guo}, \bibinfo{person}{Jianqi Zhang}, \bibinfo{person}{Changwen Zheng}, {and} \bibinfo{person}{Wenwen Qiang}.} \bibinfo{year}{2024}\natexlab{}.
\newblock \showarticletitle{Not All Frequencies Are Created Equal: Towards a Dynamic Fusion of Frequencies in Time-Series Forecasting}. In \bibinfo{booktitle}{\emph{ACM Multimedia 2024}}.
\newblock


\bibitem[Zhang and Yan(2023)]%
        {zhang2023crossformer}
\bibfield{author}{\bibinfo{person}{Yunhao Zhang} {and} \bibinfo{person}{Junchi Yan}.} \bibinfo{year}{2023}\natexlab{}.
\newblock \showarticletitle{Crossformer: Transformer Utilizing Cross-Dimension Dependency for Multivariate Time Series Forecasting}. In \bibinfo{booktitle}{\emph{The Eleventh International Conference on Learning Representations}}.
\newblock


\bibitem[Zheng et~al\mbox{.}(2023)]%
        {HybridZheng}
\bibfield{author}{\bibinfo{person}{Wendong Zheng}, \bibinfo{person}{Putian Zhao}, \bibinfo{person}{Gang Chen}, \bibinfo{person}{Huihui Zhou}, {and} \bibinfo{person}{Yonghong Tian}.} \bibinfo{year}{2023}\natexlab{}.
\newblock \showarticletitle{A Hybrid Spiking Neurons Embedded {LSTM} Network for Multivariate Time Series Learning Under Concept-Drift Environment}.
\newblock \bibinfo{journal}{\emph{{IEEE} Trans. Knowl. Data Eng.}} \bibinfo{volume}{35}, \bibinfo{number}{7} (\bibinfo{year}{2023}), \bibinfo{pages}{6561--6574}.
\newblock


\bibitem[Zhou et~al\mbox{.}(2022)]%
        {zhou2022fedformer}
\bibfield{author}{\bibinfo{person}{Tian Zhou}, \bibinfo{person}{Ziqing Ma}, \bibinfo{person}{Qingsong Wen}, \bibinfo{person}{Xue Wang}, \bibinfo{person}{Liang Sun}, {and} \bibinfo{person}{Rong Jin}.} \bibinfo{year}{2022}\natexlab{}.
\newblock \showarticletitle{{FEDformer}: Frequency enhanced decomposed transformer for long-term series forecasting}. In \bibinfo{booktitle}{\emph{Proc. 39th International Conference on Machine Learning (ICML 2022)}} (Baltimore, Maryland).
\newblock


\bibitem[Zhou et~al\mbox{.}(2023)]%
        {gpt4ts}
\bibfield{author}{\bibinfo{person}{Tian Zhou}, \bibinfo{person}{Peisong Niu}, \bibinfo{person}{Xue Wang}, \bibinfo{person}{Liang Sun}, {and} \bibinfo{person}{Rong Jin}.} \bibinfo{year}{2023}\natexlab{}.
\newblock \showarticletitle{One Fits All: Power General Time Series Analysis by Pretrained {LM}}. In \bibinfo{booktitle}{\emph{Thirty-seventh Conference on Neural Information Processing Systems}}.
\newblock


\bibitem[Zhu et~al\mbox{.}(2024)]%
        {zhu2024}
\bibfield{author}{\bibinfo{person}{Peng Zhu}, \bibinfo{person}{Yuante Li}, \bibinfo{person}{Yifan Hu}, \bibinfo{person}{Qinyuan Liu}, \bibinfo{person}{Dawei Cheng}, {and} \bibinfo{person}{Yuqi Liang}.} \bibinfo{year}{2024}\natexlab{}.
\newblock \showarticletitle{LSR-IGRU: Stock Trend Prediction Based on Long Short-Term Relationships and Improved GRU}. In \bibinfo{booktitle}{\emph{Proceedings of the 33rd ACM International Conference on Information and Knowledge Management}} (Boise, ID, USA) \emph{(\bibinfo{series}{CIKM '24})}. \bibinfo{publisher}{Association for Computing Machinery}, \bibinfo{address}{New York, NY, USA}, \bibinfo{pages}{5135–5142}.
\newblock
\showISBNx{9798400704369}
\urldef\tempurl%
\url{https://doi.org/10.1145/3627673.3680012}
\showDOI{\tempurl}


\end{thebibliography}
